\documentclass[10pt,twocolumn,letterpaper]{article}

\usepackage{iccv}              

\definecolor{iccvblue}{rgb}{0.21,0.49,0.74}
\usepackage[pagebackref,breaklinks,colorlinks,allcolors=iccvblue]{hyperref}

\usepackage{multirow}

\def\paperID{10638} 
\def\confName{ICCV}
\def\confYear{2025}

\title{Importance-Based Token Merging for Efficient Image and Video Generation}

\author{
Haoyu Wu\textsuperscript{1} \qquad
Jingyi Xu\textsuperscript{1} \qquad
Hieu Le\textsuperscript{2} \qquad
Dimitris Samaras\textsuperscript{1} \\
\textsuperscript{1}Stony Brook University \qquad
\textsuperscript{2}EPFL\\
}

\begin{document}
\maketitle
\newcommand{\HL}[1]{{\color{orange}{\bf HL: #1}}} 
\newcommand{\hl}[1]{{\color{orange} #1}}
\newcommand{\h}[1]{{\color{blue} #1}}

\begin{abstract}

Token merging can effectively accelerate various vision systems by processing groups of similar tokens only once and sharing the results across them. However, existing token grouping methods are often ad hoc and random, disregarding the actual content of the samples. 
We show that preserving high-information tokens during merging—those essential for semantic fidelity and structural details—significantly improves sample quality, producing finer details and more coherent, realistic generations. Despite being simple and intuitive, this approach remains underexplored.

To do so, we propose an importance-based token merging method that prioritizes the most critical tokens in computational resource allocation, leveraging readily available importance scores, such as those from classifier-free guidance in diffusion models. Experiments show that our approach significantly outperforms baseline methods across multiple applications, including text-to-image synthesis, multi-view image generation, and video generation with various model architectures such as Stable Diffusion, Zero123++, AnimateDiff, or PixArt-$\alpha$.
\end{abstract}
\section{Introduction}
\label{sec:intro}

The rise of powerful diffusion models such as DALL-E~\cite{Ramesh2022HierarchicalTI}, Stable Diffusion (SD)~\cite{rombach2022high}, or Imagen~\cite{saharia2022photorealistic} has dramatically changed the landscape of generative AI~\cite{ho2020denoising, song2020denoising,rombach2022high, saharia2022photorealistic, blattmann2023stable, guo2023animatediff}. At their core, these models operate by iteratively denoising through multiple passes of a backbone network, processing a substantial number of tokens, which are particularly computationally intensive. To reduce the computational demands, prior work~\cite{Bolya2022TokenMY, Bolya2023TokenMF, kim2024token} has explored merging similar tokens and sharing the results across member tokens.

\begin{figure}[t]
  \centering
    Prompt: ``\textit{A cute owl wearing a wizard hat.}''
  \includegraphics[width=1.0\linewidth]{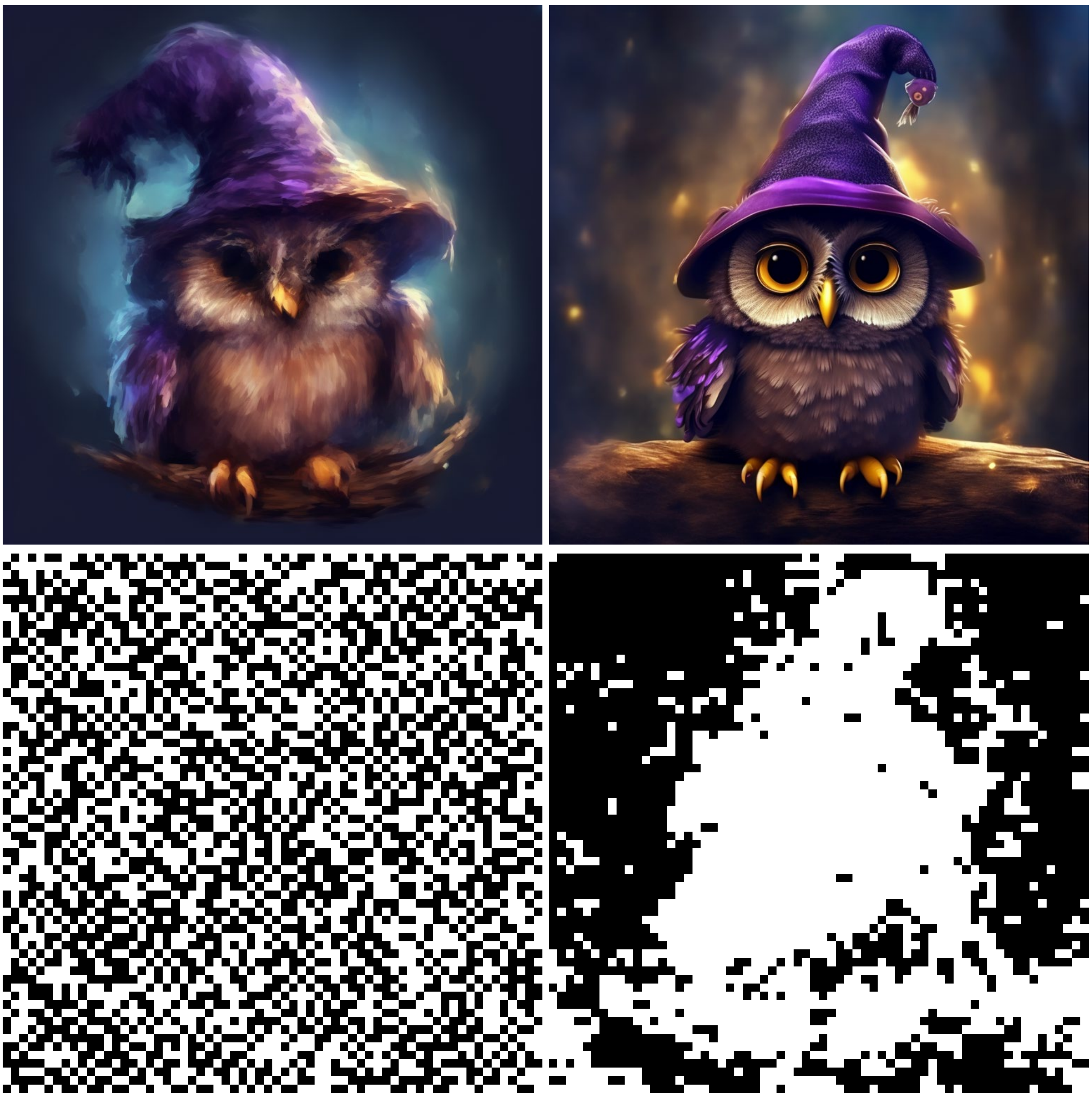}
    \makebox[0.5\linewidth]{\centering (a) Spatial~\cite{Bolya2023TokenMF}}%
    \makebox[0.5\linewidth]{\centering (b) Importance-based (Ours)}
  \vspace{-4mm}
  \caption{\textbf{Importance-based Token Merging.} Our method prioritizes important tokens during token merging, resulting in images with greater details in essential areas compared to ToMeSD~\cite{Bolya2023TokenMF}. In the second row, we show the regions (in white) where computation (\eg, attention) will take place after token merging.}
  \label{fig:teaser}
  \vspace{-3mm}
\end{figure}

However, such a merging process can also significantly degrade image quality, losing important details and structure. This typically happens when merging occurs in highly informative image regions, where critical visual details are compressed or discarded. One major reason is that existing methods rely on ad hoc heuristics for grouping tokens, often based on spatial proximity or fixed patterns without considering their content.  
Thus, destination tokens, where the other tokens will merge into, are chosen randomly within predefined regions, sometimes pushing important visual elements into less relevant areas.
This leads to suboptimal resource allocation, where crucial textures, edges, or fine-grained structures are lost, while less important regions retain more detail than necessary, as can be seen in~\cref{fig:teaser} (a). Existing methods rarely consider token content explicitly when merging.

To address this limitation, we propose a novel token selection method that prioritizes computational resources in areas of high visual and semantic importance. Unlike existing approaches that rely on random or fixed-pattern selection, our method ensures that merging decisions are guided by actual content relevance. 
Rather than treating all tokens equally, we use per-token importance signals to identify key regions and ensure that merging prioritizes preserving more important tokens. Note that there are many proxies for importance, such as attention scores, saliency maps, or user-provided bounding boxes, all of which can be integrated into our merging method. Specifically in our case, we point out that an excellent choice is classifier-free guidance (CFG)~\cite{ho2022classifier}, as it inherently highlights regions that strongly influence model outputs and is obtainable with no additional cost.


More specifically, instead of merging tokens arbitrarily across the image, we first construct a pool of high-importance tokens. Then, token partitioning is only performed within this pool using a soft-matching strategy. This ensures that the most relevant tokens serve as anchors for merging, preserving key details while also maintaining diversity among the anchor tokens. Doing so facilitates a more efficient allocation of computational resources, preventing redundancy and ensuring that merging decisions are both meaningful and effective. Compared to simply selecting the top-k important tokens—which can lead to redundancy and suboptimal token assignments—our method strikes a balance between relevance and diversity, resulting in more coherent and detailed generations.

We apply our token merging strategy across three key applications: text-to-image generation, multi-view generation, and text-to-video generation, as well as across various model structures such as U-Net or transformers.
Compared to baseline methods, our method consistently demonstrates superior performance, delivering results with higher fidelity and significantly improved image details across all tested scenarios.

Our contributions are as follows:
\begin{itemize}
\item We propose a novel importance-based token merging paradigm for diffusion models, designed to preserve crucial image content. The importance score can be easily obtained from CFG at no additional cost.

\item We design a novel token-partitioning strategy based on a pool of important tokens to improve generation quality.

\item Our method achieves state-of-the-art performance across various diffusion model tasks, including text-to-image synthesis, multi-view generation, and video generation, as well as across model architectures including U-Net and transformers.
\end{itemize}
\section{Related Work}
\label{sec:related_work}

\subsection{Diffusion Models}
\label{ssec:diffusion}
Diffusion models~\cite{sohl2015deep, dhariwal2021diffusion, ho2020denoising, song2020denoising} are a class of generative models that iteratively transform random noise into complex data structures, such as images, by gradually reversing a diffusion process. In these models, data is progressively corrupted by adding noise, and the model learns to recover the original data through iterative denoising with learned parameters. This approach has demonstrated impressive results in generating high-quality, realistic images~\cite{rombach2022high, saharia2022photorealistic, peebles2023scalable, chen2023pixartalpha, Ramesh2022HierarchicalTI, nichol2021glide, graikos2024learned, meng2021sdedit, xue2024raphael, balaji2022ediff, karras2025guiding}, videos~\cite{singer2022make, blattmann2023stable, guo2023animatediff, opensora, pku_yuan_lab_and_tuzhan_ai_etc_2024_10948109, brooks2024video}, 3D content~\cite{poole2022dreamfusion, liu2023zero, long2024wonder3d, shi2023zero123++, li20223ddesigner, luo2021diffusion, nichol2022point}, and audio~\cite{kong2020diffwave, huang2023make}.
A popular diffusion model architecture is introduced by Stable Diffusion~\cite{rombach2022high}, which uses a U-Net with transformer blocks. It first encodes a noisy image as latent tokens, which are processed through transformer blocks comprising self-attention, MLP, and cross-attention layers. This design has been extended for multi-view generation with multi-view attention~\cite{shi2023zero123++} and for video generation with temporal layers~\cite{guo2023animatediff}. More recent approaches~\cite{chen2023pixartalpha, opensora} replace the U-Net with a fully transformer-based architecture for improved scalability.

In diffusion models, classifier-free guidance (CFG)~\cite{ho2022classifier} is a technique that enhances fidelity and detail in generation by guiding the model toward conditioned inputs without the need for an external classifier. 
Recent work~\cite{zhao2023magicfusion, wang2024high} suggests a connection between CFG and saliency. Our study advances this by identifying CFG as an indicator of token importance, enabling more effective token merging.

To reduce the cost of diffusion inferences, various techniques have been proposed. Some approaches require retraining, including better model
structures~\cite{rombach2022high, pernias2023wurstchen, kim2023architectural}, model pruning~\cite{fang2023structural}, model compression~\cite{zhao2023mobilediffusion, yang2023diffusion, li2024snapfusion}, step distillation~\cite{salimans2022progressive, meng2023distillation, sauer2025adversarial, liu2023instaflow, habibian2024clockwork}, and consistency regularization~\cite{song2023consistency, luo2023latent}. Others avoid retraining, such as improved sampling to reduce inference steps~\cite{song2020denoising, liu2022pseudo, lu2022dpm, lu2022dpm_pp}, caching~\cite{ma2024deepcache, wimbauer2024cache, chen2024delta, zhao2024real, kahatapitiya2024adaptive, so2023frdiff, lv2024fastercache, li2023faster}, model quantization~\cite{li2023q, chen2024q, he2024ptqd, wang2024quest, deng2024vq4dit, so2024temporal}, and token reduction~\cite{kahatapitiya2025object, Bolya2023TokenMF, Bolya2022TokenMY}. In this work, we focus on improving token reduction. Notably, our method is compatible with other diffusion acceleration techniques, enabling combined use for further speedup.

\subsection{Token Reduction}
\label{ssec:token_reduction}
Token reduction~\cite{haurum2023tokens, Bolya2022TokenMY, wang2024zero} decreases the number of tokens to process by pruning or merging them. It is widely applied in tasks such as classification~\cite{marin2023token, Bolya2022TokenMY, haurum2023tokens, liang2022not}, segmentation~\cite{xu2022groupvit, kienzle2024segformer++}, detection~\cite{liu2024revisiting}, video understanding~\cite{choi2024vid}, and large language models~\cite{li2024tokenpacker, jin2024chat, song2024less}. Token pruning methods include removing tokens based on attention scores~\cite{fayyaz2022adaptive, haurum2023tokens, liang2022not, long2023beyond, wei2023joint, wu2023ppt, xu2022evo}, using the Gumbel-Softmax trick for selective pruning~\cite{jang2016categorical, kong2022spvit, liu2024revisiting, rao2021dynamicvit, wei2023joint}, developing sampling methods~\cite{fayyaz2022adaptive, yin2022vit}, integrating sparsity in the model~\cite{chen2021chasing, li2022sait}, and reinforcement learning-based methods~\cite{pan2023interpretability}. For token merging~\cite{Bolya2022TokenMY, haurum2022multi, renggli2022learning, xu2022groupvit, zong2022self, tran2024accelerating, wang2024efficient, leelearning}, approaches include bipartite soft matching~\cite{Bolya2022TokenMY, Bolya2023TokenMF}, K-Means~\cite{marin2023token}, K-Medoids clustering~\cite{marin2023token}, and Density-Peak Clustering with K-Nearest Neighbors (DPC-KNN)~\cite{zeng2022not}. Additionally, soft merging methods assign tokens to multiple clusters before merging them~\cite{haurum2022multi, renggli2022learning, zong2022self}.

Recent studies have applied token reduction to diffusion model inference~\cite{Bolya2023TokenMF, kim2024token, haurum2025agglomerative, wang2024attention, li2024vidtome, kahatapitiya2025object, zhao2024dynamic, liang2024looking, wang2024sparsedm, smith2024todo, lutoma}. ToMeSD~\cite{Bolya2023TokenMF} introduced token merging to Stable Diffusion~\cite{rombach2022high} with a training-free method. 
The typical token merging procedure for diffusion models selects destination tokens from input feature tokens and utilizes bipartite soft matching to merge redundant tokens.
The reduced token set is processed by operations like attention, and is then copied back to the merged token locations, ensuring the final token count matches the input. 
Although this token merging approach performs well, its selection of destination tokens is not optimal, \ie, spatially random across the image.
ToFu~\cite{kim2024token} combines token pruning and merging but still follows similarly suboptimal destination token selection strategy.
In our work, we propose that selecting destination tokens based on their importance improves the fidelity and quality in generation, and this importance can be easily obtained via classifier-free guidance without additional cost. AT-EDM~\cite{wang2024attention} uses self-attention maps to derive token importance, but this requires first performing full self-attention and increases peak memory usage. ATC~\cite{haurum2025agglomerative} applies bottom-up hierarchical clustering for better token merging, but we show it is costly for generative tasks.

\section{Method}
\label{sec:method}

\begin{figure}[t]
\centerline{\includegraphics[width=0.5\textwidth]{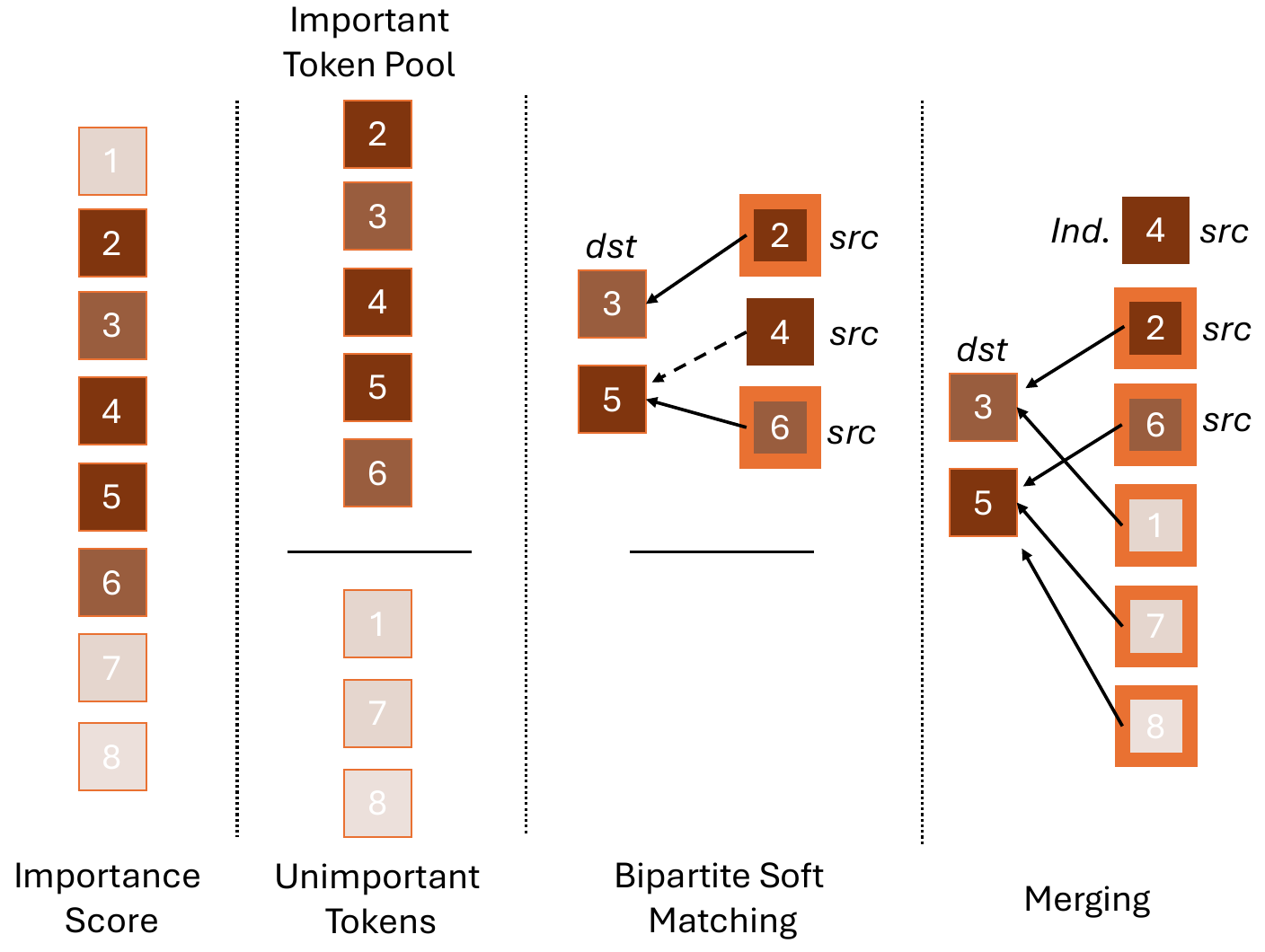}}
\vspace{-1mm}
\caption{\textbf{Overview.} We propose an importance-based token merging method. 
The importance of each token can be determined using classifier-free guidance.
These scores, visualized with colors ranging from light to dark (indicating less to more important tokens), are used to construct a pool of important tokens. We randomly select a set of destination (\textit{dst}) tokens from this pool and the remaining important tokens become source (\textit{src}) tokens. Bipartite soft matching is then performed between the \textit{dst} tokens and \textit{src} tokens. \textit{src} tokens without a suitable match are considered independent tokens (\textit{ind.}). All other \textit{src} tokens and unimportant tokens are merged with the destination tokens for subsequent computational steps.}
\label{fig:main}
\vspace{-1mm}
\end{figure}

We propose a novel importance-based token-merging method, summarized in \cref{fig:main}, allowing for a more efficient allocation of computing resources. We highlight that classifier-free guidance serves as an effective importance indicator. Further, we propose using a dynamic pool of important tokens, where the pool size adapts to the token merging ratio, optimizing resource allocation and reducing redundancy.

\paragraph{Important Tokens.}

\begin{figure}[t]
\begin{center}
``\textit{Fresh citrus and berries}''
\includegraphics[width=1.0\linewidth]{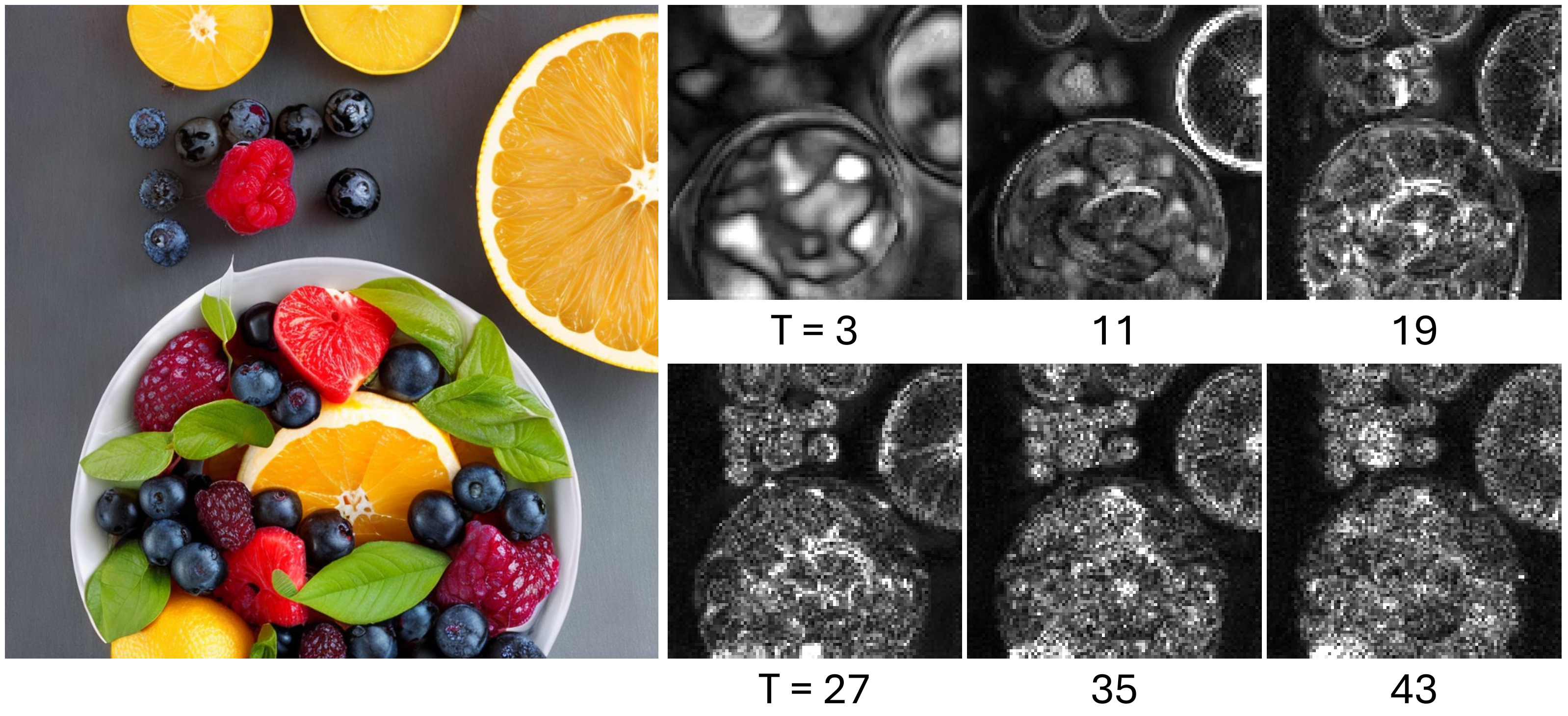}
\end{center}
\vspace{-6mm}
\caption{\textbf{Importance Maps.} We present token importance maps derived from classifier-free guidance (CFG) across diffusion inference timesteps. These maps highlight areas significantly align with the user prompt. In the early steps, they capture the semantics and structure of the image relevant to the prompt, while in later steps, they focus on finer details of the objects the user intends to generate. The generated image is shown on the left for reference.}
\label{fig:cfg_map}
\vspace{-3mm}
\end{figure}

Selecting which tokens to serve as anchors (destination tokens) for merging is critical for generating high-quality content. This is because these tokens correspond to the primary image regions where subsequent computations, such as attention, are applied.
In principle, all reliable per token importance signals, such as cross attention maps, user provided bounding boxes, or saliency maps, can be integrated with our method. We find that classifier-free guidance (CFG)~\cite{ho2022classifier} serves as an excellent indicator. 
CFG modifies the noise prediction to improve sample control. It is designed to steer the predicted noise in a direction more aligned with the condition: 
\begin{equation}
\tilde{\epsilon}_\theta(\mathbf{x}_t, t) = \epsilon_\theta(\mathbf{x}_t, t) + w \cdot (\epsilon_\theta(\mathbf{x}_t \mid y, t) - \epsilon_\theta(\mathbf{x}_t, t)),
\label{eq:classifier_free_guidance}
\end{equation}
where $\epsilon_\theta(\mathbf{x}_t \mid y, t)$ is the noise prediction conditioned on input $y$, $\epsilon_\theta(\mathbf{x}_t, t)$ is the unconditional noise prediction, and $w$ is the guidance weight.
It is widely used in diffusion models and incurs no additional computational cost. For each token, we calculate its importance as the absolute value of its CFG score:
\begin{equation}
\begin{aligned}
\text{importance} &= \left| \epsilon_\theta(\mathbf{x}_t \mid y, t) - \epsilon_\theta(\mathbf{x}_t, t) \right| \\
                   &\approx \left| -\sigma_t \nabla_{\mathbf{x}_t} \log p(y \mid \mathbf{x}_t) \right|,
\end{aligned}
\label{eq:importance}
\end{equation}
where $\sigma_t$ is the noise scale. The guidance term effectively estimates the gradient of the log-likelihood of the conditioning variable $y$ (e.g., a text prompt) with respect to the noisy sample $\mathbf{x}_t$~\cite{ho2022classifier}. Thus, the CFG magnitude can be interpreted as a saliency or importance measure: tokens with a high CFG magnitude have stronger influences in steering the generation toward satisfying the condition $y$. In~\cref{fig:cfg_map}, we provide an example of the resulting importance maps.


\begin{figure}[t]
\begin{center}
``\textit{A delicate pink rose in full bloom, detailed petals.}''
\includegraphics[width=1.0\linewidth]{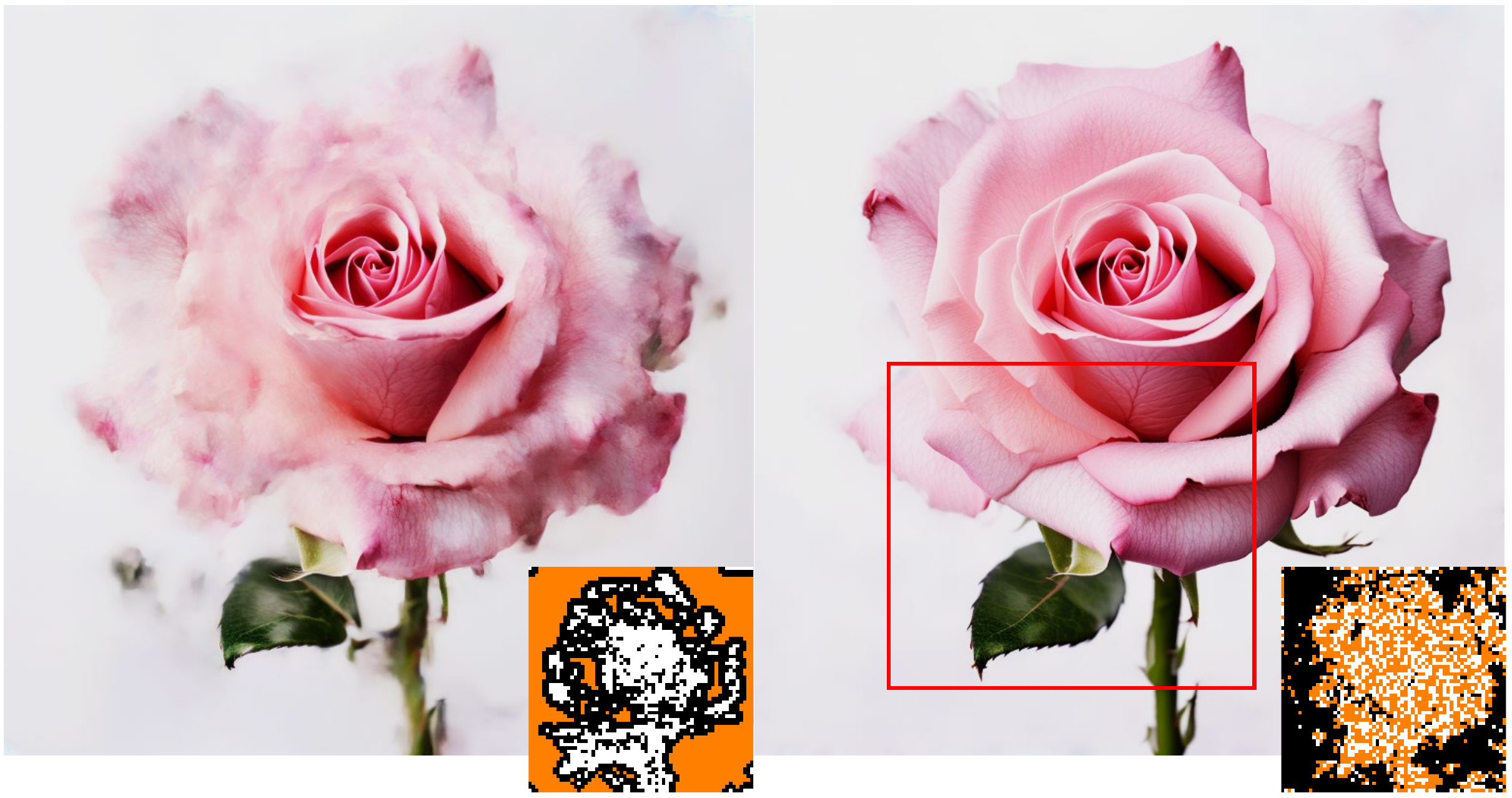}
\makebox[0.5\linewidth]{\centering (a) Top-k}%
\makebox[0.5\linewidth]{\centering (b) Ours}
\end{center}
\vspace{-6mm}
\caption{We compare our method with an approach that uses the top-k important tokens as destination tokens (\textit{dst}) for token merging. The computation locations after token merging are illustrated as non-black pixels in the bottom-left windows. They include locations of \textit{dst} tokens, which are shown in white, and independent tokens (some other tokens that lack a similar \textit{dst} token for merging), which are shown in orange. Our method produces more structured and detailed image, as highlighted in the red box.}
\label{fig:topk}
\vspace{-4mm}
\end{figure}

\paragraph{Importance-based Token Merging.}
With the token importance scores, a naive approach is to pick the top-k tokens as destination tokens (\textit{dst})  and merge the rest tokens that are similar to them. However, this approach produces low-quality outputs due to merging inefficiency, as shown in \cref{fig:topk} (a). 
More specifically, there are two particular issues about this: 
\begin{enumerate}
    \item \textbf{Redundancy}. The top-k tokens can be very similar - but all important tokens, leading to redundancy and less intra-variant among the selected destination tokens.
    \item \textbf{Unimportant Independent Tokens}. In the token merging pipeline, some tokens lack similar destination tokens, making them ``independent'' and remain unmerged. In the top-k approach, background or irrelevant tokens often become independent due to the absence of suitable matches among the important tokens, as shown in~\cref{fig:topk}.
\end{enumerate}

To avoid these issues, we propose to first create a pool of the most important tokens and then, ensure that both destination tokens and independent tokens are drawn from this set. To do so, destination tokens are randomly sampled from the pool, while independent tokens are selected as those in the important token pool that are most dissimilar to the chosen destination tokens. 
This approach ensures that all computations following the merging step operate on important tokens, leading to improved detail preservation, as illustrated in \cref{fig:topk} (b).

To determine the optimal pool size, we adapt it based on the token merging ratio $r$. Specifically, we set the pool size as $\mathbf{P}=(1 - r) \cdot (1 + p)$, where $p$ is a hyper-parameter of our method. With a constrained token processing budget, \ie, a high token merging ratio $r$, the pool size remains small, ensuring the selection of only the most critical tokens. Conversely, with a larger compute budget, the pool size increases, reducing the likelihood of selecting redundant tokens as destination tokens. 

\begin{figure*}[t]
\begin{center}
\includegraphics[width=0.88\textwidth]{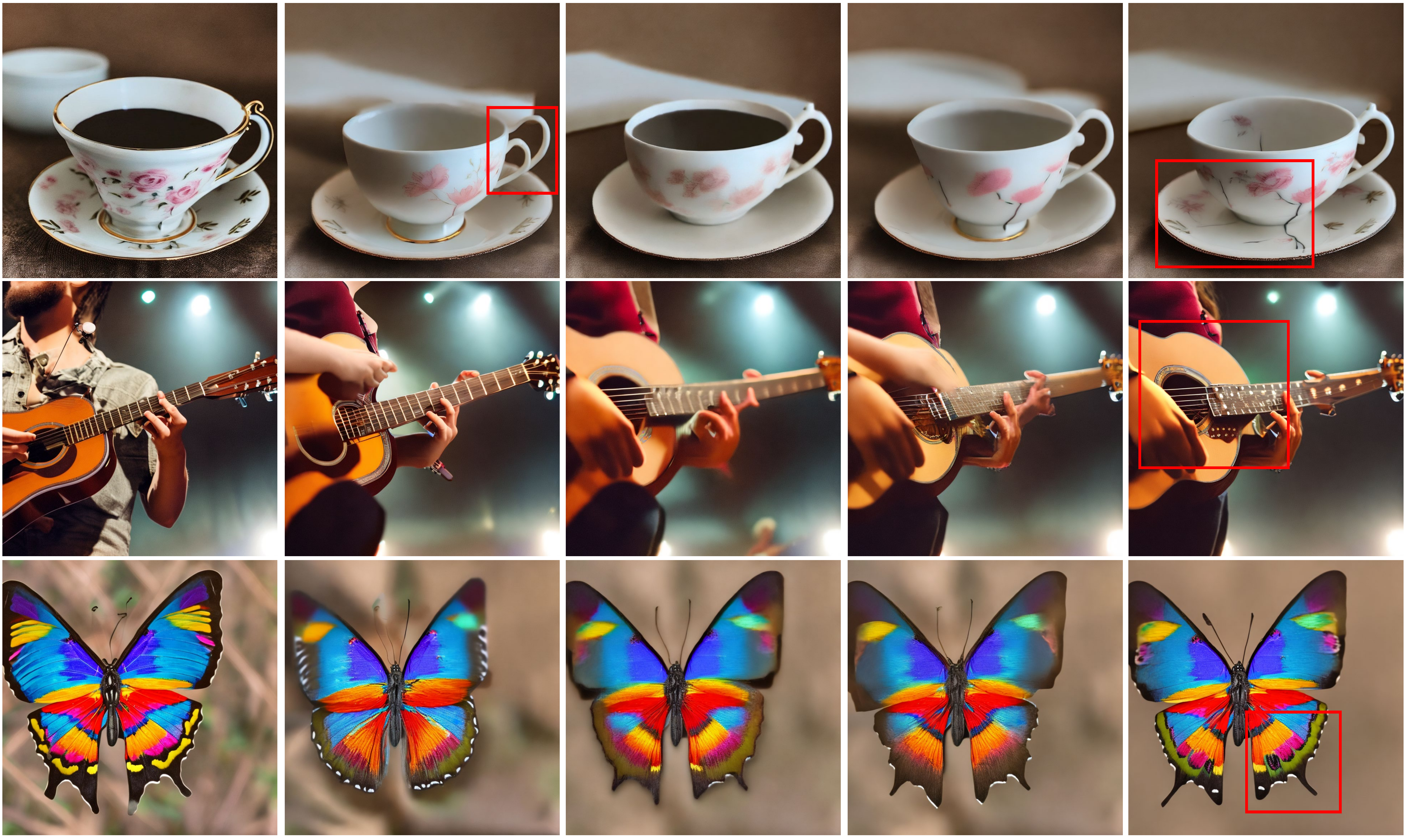}
\makebox[0.176\linewidth]{\centering (a) SD~\cite{rombach2022high} $\sim$
 8s}%
\makebox[0.176\linewidth]{\centering (b) ATC~\cite{haurum2025agglomerative} $\sim$ mins}%
\makebox[0.176\linewidth]{\centering (c) ToFu~\cite{kim2024token} $\sim$ 5s}%
\makebox[0.176\linewidth]{\centering (d) ToMe.~\cite{Bolya2023TokenMF} $\sim$ 5s}%
\makebox[0.176\linewidth]{\centering (e) Ours $\sim$ 5s}
\end{center}
\vspace{-6mm}
\caption{\textbf{Qualitative comparison of text-to-image generation.} The first column shows results from Stable Diffusion (SD)~\cite{rombach2022high}, while the subsequent columns show SD combined with various token merging methods.
As highlighted in red boxes, our approach consistently produces finer details with coherent structures. Note that ATC requires minutes to generate an image, whereas other methods, including ours, complete the task in seconds. The token merging ratio is 0.7. Please see the supplementary for prompts. Best viewed with zoom-in.}
\label{fig:t2i}
\vspace{-6mm}
\end{figure*}

\paragraph{Token Merging in Diffusion Inference.}
At time-step $t$ of the diffusion inference, a diffusion model layer, \eg a transformer layer, takes $\mathbf{N}$ tokens as input. 
The token merging ratio is $r$.
Based on the token importance derived from the previous timestep's classifier-free guidance, we select the top $\mathbf{K} = \mathbf{N} \cdot (1 - r) \cdot (1 + p)$ tokens as the important token pool, denoted as $\mathbf{A}$. From $\mathbf{A}$, we randomly pick $\mathbf{D} = \mathbf{N} \cdot k$ tokens to form the destination (\textit{dst}) set. Here, $p$, $k$ are hyper-parameters.
The remaining tokens in $\mathbf{A}$ become the source (\textit{src}) set. 

Next, we perform bipartite soft matching by computing pairwise cosine similarities between \textit{src} and \textit{dst} tokens. Each \textit{src} token is linked to its most similar \textit{dst}. 
Then, in \textit{src} set, we select the top
$\mathbf{I} = \mathbf{N}\cdot(1 - k  - r)$ tokens
with the smallest similarity to their closest \textit{dst} tokens to serve as independent tokens. The remaining \textit{src} tokens and unimportant tokens are merged into their corresponding \textit{dst} tokens.
The merging is performed via averaging all grouped tokens.
After merging, the number of tokens reduces to $\mathbf{I}+\mathbf{D} = \mathbf{N}  \cdot (1 - r)$ tokens.
The diffusion model layer then processes this reduced token set, the merged locations are filled with the corresponding processed \textit{dst} tokens, maintaining the same output shape as the input.

\section{Experiments}
\label{sec:exp}

\subsection{Experimental Settings}
\label{ssec:exp_setting}


\paragraph{Text-to-image Generation.} We use Stable Diffusion 2 (SD)~\cite{rombach2022high} as the base model, a text-to-image latent diffusion model with a U-Net architecture that generates 768x768 images. Our token merging method is compared to ToFu~\cite{kim2024token}, ToMeSD~\cite{Bolya2023TokenMF} and ATC~\cite{haurum2025agglomerative}. Due to ATC’s slow inference (several minutes per image), we only display its visual results. For quantitative comparison, we follow previous studies~\cite{saharia2022photorealistic, xue2024raphael, balaji2022ediff, Ramesh2022HierarchicalTI} and report FID~\cite{heusel2017gans} and CLIP scores~\cite{radford2021learning, hessel2021clipscore} for zero-shot image generation on the MS-COCO 2014 validation dataset~\cite{lin2014microsoft}, with 30K randomly sampled image-caption pairs.

We also experiment with a diffusion transformer, PixArt-$\alpha$~\cite{chen2023pixartalpha}, for text-to-image generation. We compare our method with ToMeSD and similarly evaluate on the MS-COCO dataset. Unless otherwise stated, ``text-to-image'' in this paper refers to generation based on Stable Diffusion.

\paragraph{Multi-view Diffusion.} We use Zero123++ v1.2~\cite{shi2023zero123++} as the base model. Zero123++ is an image-conditioned multi-view latent diffusion model that generates six novel views at a resolution of 320×320. We compare our method to ToMeSD. Following evaluation protocols of prior work~\cite{liu2023syncdreamer, long2024wonder3d}, we test on GSO dataset~\cite{downs2022google}, which comprises 30 everyday objects, and compute PSNR, SSIM~\cite{wang2004image}, and LPIPS metrics~\cite{zhang2018unreasonable} to evaluate the similarity between the generated images and ground truth. We also include visual comparisons using in-the-wild images as input.

\begin{figure}[t]
\begin{center}
\includegraphics[width=1\linewidth]{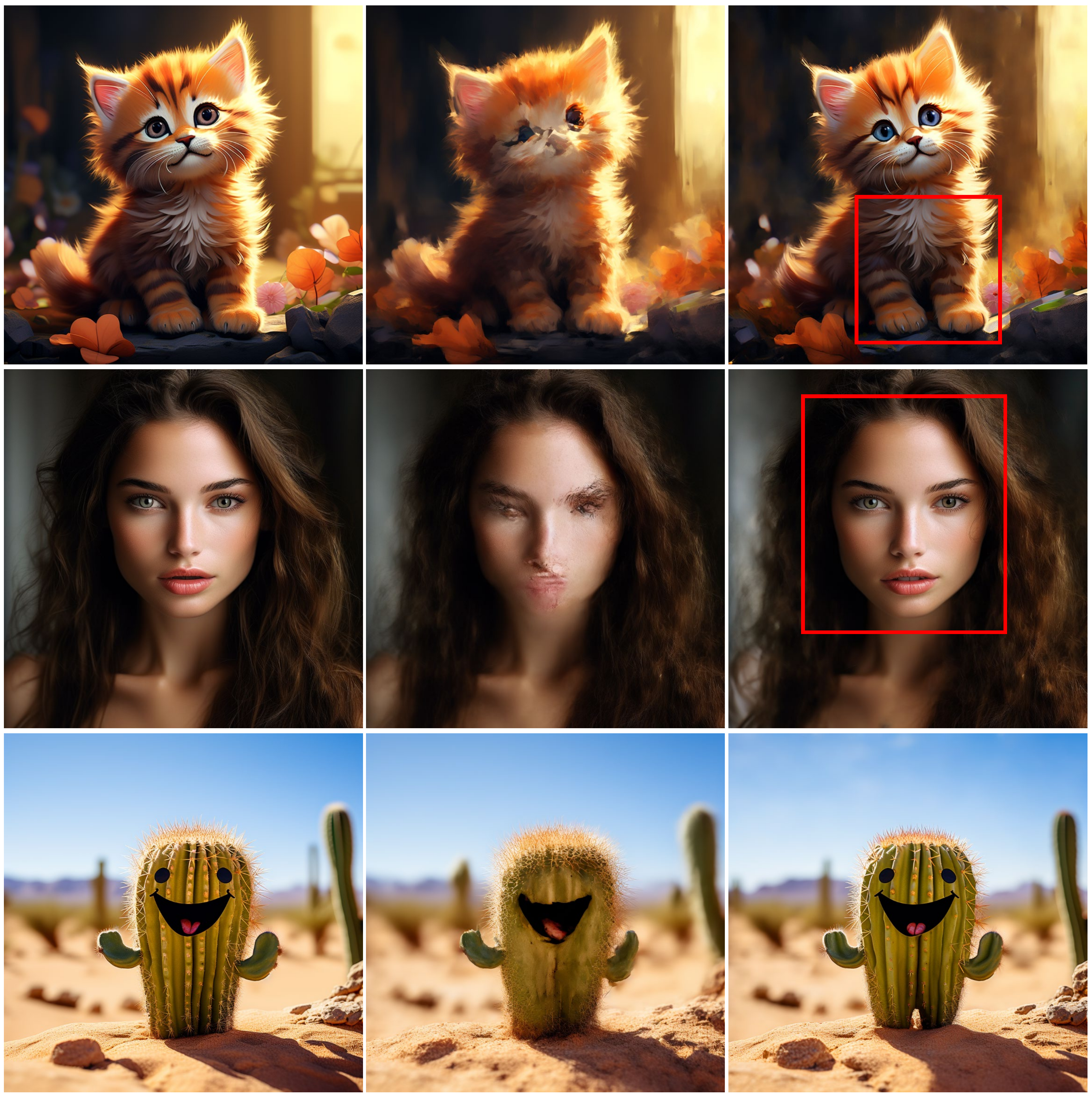}
\makebox[0.33\linewidth]{\centering (a) PixArt-$\alpha$~\cite{chen2023pixartalpha}}%
\makebox[0.33\linewidth]{\centering (b) ToMeSD~\cite{Bolya2023TokenMF}}%
\makebox[0.33\linewidth]{\centering (c) Ours}
\end{center}
\vspace{-6mm}
\caption{\textbf{Token merging for diffusion transformer.} We apply ToMeSD~\cite{Bolya2023TokenMF} and our method to PixArt-$\alpha$~\cite{chen2023pixartalpha} with a merging ratio of 0.3. Detailed generations are highlighted with red boxes. Best viewed with zoom-in. Please see the supplementary for prompts.}
\label{fig:pixart}
\vspace{-6mm}
\end{figure}

\begin{figure*}[t]
\begin{center}
\includegraphics[width=1\textwidth]{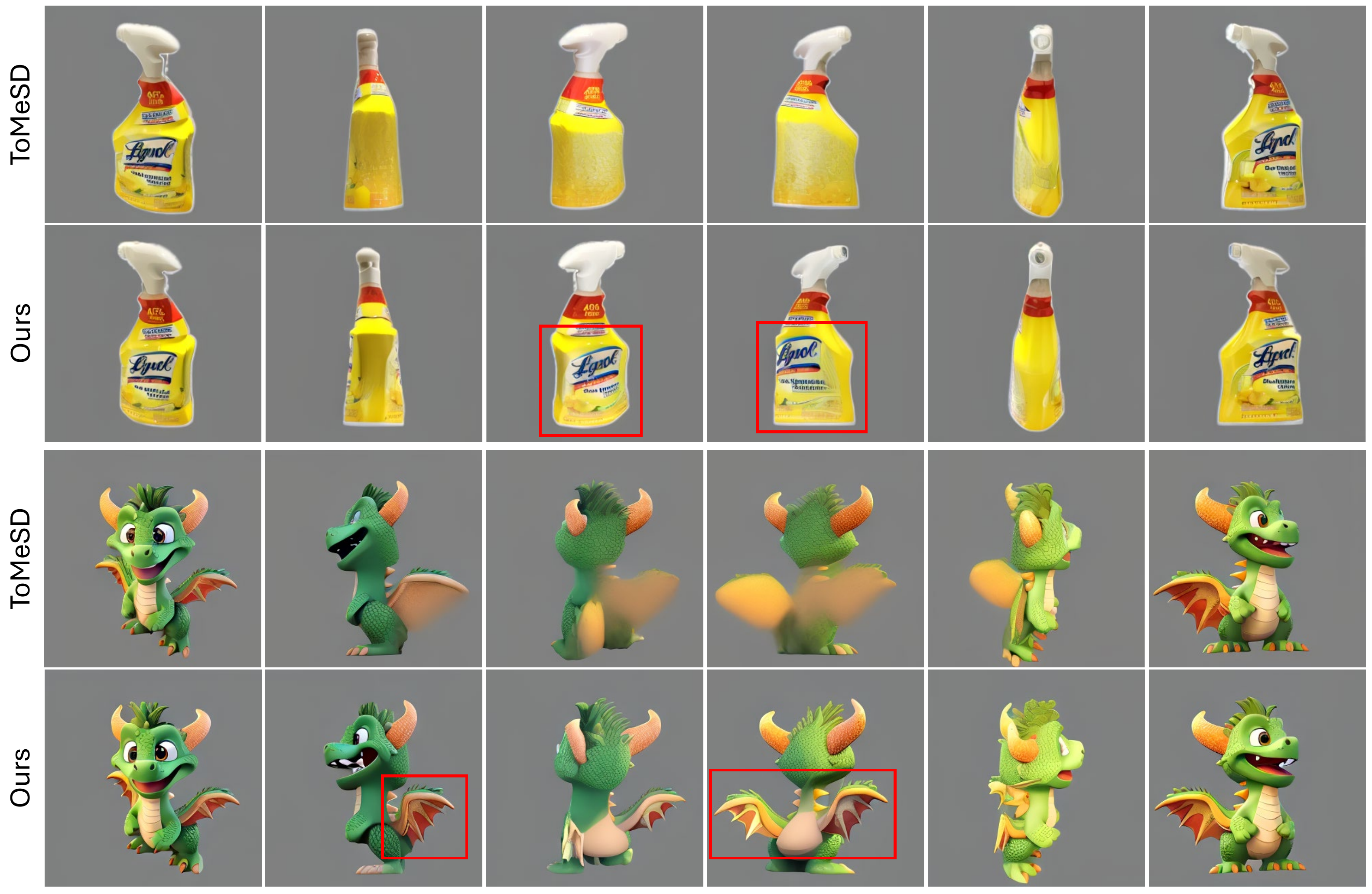}
\end{center}
\vspace{-6mm}
\caption{\textbf{Qualitative comparison of multi-view diffusion.} We apply ToMeSD~\cite{Bolya2023TokenMF} and our token merging method to the multi-view diffusion model. We use Zero123++~\cite{shi2023zero123++} as the base model and a merging ratio of 0.6. Our method outputs finer details, as highlighted in red boxes. Best viewed with zoom-in. Please refer to the supplementary for input images.}
\label{fig:mv}
\vspace{-3mm}
\end{figure*}

\paragraph{Video Diffusion.} We adopt AnimateDiff v3~\cite{guo2023animatediff} as the base model, which adds temporal attention layers to turn the text-to-image model, \ie Stable Diffusion, into a video diffusion model. This model generates 16-frame videos at a resolution of 512×512. We compare our token merging method with ToMeSD and evaluate using VBench~\cite{huang2024vbench}.
We report the semantic, quality, and total scores from VBench. The semantic score assesses alignment between the generated videos and the user prompt, focusing on entity types, attributes, and styles. The quality score evaluates the temporal consistency and visual quality of the generated videos.

\paragraph{Other Metrics.} We report TFLOPs, latency, and GPU memory usage, with and without memory-efficient attention~\cite{xFormers2022}, on an NVIDIA A5000 GPU. Inference uses float16 precision. We estimate TFLOPs for a single diffusion sampling step. For latent diffusion models, the inference cost is measured exclusively in the latent space.

\paragraph{Implementation Details.} For text-to-image synthesis with Stable Diffusion~\cite{rombach2022high}, we merge tokens in the self-attention layers of the first and last model blocks, similar to ToMeSD~\cite{Bolya2023TokenMF}. 
For PixArt-$\alpha$~\cite{chen2023pixartalpha}, we apply token merging to self-attention and cross-attention in the middle half of the model layers (7–20 out of 28) for simplicity.
For multi-view diffusion, we merge tokens in the self-attention layers of the first two and last two blocks of Zero123++~\cite{shi2023zero123++}. 
For video diffusion, we merge tokens in the first and last blocks of AnimateDiff~\cite{guo2023animatediff}. 
The hyper-parameter $p$, which determines the size of our important token pool, is set to 0.4, 0.6, and 0.8 for image, multi-view, and video generation tasks, respectively. The number of destination tokens ($k$) remains consistent across all tasks and follows the setting used in ToMeSD, utilizing 25\% of the total tokens.

\subsection{Results}
\label{ssec:results}

In~\cref{tab:t2i} and~\cref{fig:t2i}, we compare different token merging methods applied to Stable Diffusion 2~\cite{rombach2022high} for text-to-image generation. Our method consistently outperforms baselines, especially at higher token merging ratios ($r$). For example, at $r=0.75$, our method achieves an FID of 17.75 versus ToMeSD~\cite{Bolya2023TokenMF}'s 20.89. Agglomerative Token Clustering (ATC)~\cite{haurum2025agglomerative} is not included due to its prohibitive computational cost, as it is CPU-bound and non-batched, making it impractical for large-scale evaluations. 
Notably, our method also performs well when used with cross-attention maps (\cref{ssec:ablation}).
When the merging ratio is small, such as 0.3, our important token pool becomes the whole token set, making it functionally equivalent to ToMeSD. 

\begin{table}[t]
\centering
\begin{tabular}{ccccccc}
\toprule
 \multirow{2}{*}{$r$} & \multicolumn{3}{c}{FID $\downarrow$} & \multicolumn{3}{c}{CLIP $\uparrow$} \\
\cmidrule(lr){2-4} \cmidrule(lr){5-7}
& ToFu & ToMe. & Ours & ToFu & ToMe. & Ours \\
\midrule
0    & -     &   -   & 11.88 &  -    &   -   & 31.83 \\
0.30 & 13.48 & 12.20 & 12.20 & 31.82 & 31.82 & 31.82 \\
0.50 & 15.81 & 13.50 & \textbf{13.42} & 31.81 & 31.79 & \textbf{31.83} \\
0.60 & 17.32 & 14.81 & \textbf{14.51} & 31.81  & 31.80 & \textbf{31.81} \\
0.70 & 18.94 & 17.46 & \textbf{16.22} & 31.78 & 31.78 & \textbf{31.79} \\
0.75 & 19.29 & 20.89 & \textbf{17.75} & 31.76 & 31.71 & \textbf{31.76} \\
\bottomrule
\end{tabular}
\vspace{-2mm}
\caption{\textbf{Text-to-image generation.} We apply ToFu~\cite{kim2024token}, ToMeSD~\cite{Bolya2023TokenMF} and our token merging method to Stable Diffusion~\cite{rombach2022high} across various token merging ratios $r$.}
\label{tab:t2i}
\vspace{-1mm}
\end{table}

\begin{table}[t]
\centering
\begin{tabular}{cccccc}
\toprule
 \multirow{2}{*}{$r$} & \multicolumn{2}{c}{FID $\downarrow$} & \multicolumn{2}{c}{CLIP $\uparrow$} & \multirow{2}{*}{Latency (s) $\downarrow$} \\
 \cmidrule(lr){2-3} \cmidrule(lr){4-5}
& ToMe. & Ours & ToMe. & Ours &  \\
\midrule
0    & \multicolumn{2}{c}{27.51} & \multicolumn{2}{c}{31.30} & 9.14  \\
0.3 & 34.23 & \textbf{28.50} & 30.76 & \textbf{31.05} & 8.96 \\
0.5 & 65.46 & \textbf{34.02} & 29.85 & \textbf{30.68} & 7.54 \\
0.7 & 95.46 & \textbf{50.76} & 28.67 & \textbf{29.93} & 7.14 \\
\bottomrule
\end{tabular}
\vspace{-2mm}
\caption{Comparison of our method with ToMeSD~\cite{Bolya2023TokenMF} when applied to the diffusion transformer PixArt-$\alpha$~\cite{chen2023pixartalpha}.}
\label{tab:dit}
\vspace{-1mm}
\end{table}

\begin{table}[t]
\centering
\begin{tabular}{ccccccc}
\toprule
 \multirow{2}{*}{$r$} & \multicolumn{2}{c}{PSNR $\uparrow$} & \multicolumn{2}{c}{SSIM $\uparrow$} & \multicolumn{2}{c}{LPIPS $\downarrow$} \\
\cmidrule(lr){2-3} \cmidrule(lr){4-5} \cmidrule(lr){6-7}
  & ToMe. & Ours & ToMe. & Ours & ToMe. & Ours \\
\midrule
0.40 & 14.80 & \textbf{14.82} & 0.775 & \textbf{0.777} & 0.260 & \textbf{0.259} \\
0.60 & 14.71 & \textbf{14.85} & 0.782 & \textbf{0.783} & 0.272 & \textbf{0.263} \\
0.70 & 14.18 & \textbf{14.80} & \textbf{0.787} & 0.785 & 0.302 & \textbf{0.274} \\
0.75 & 13.12 & \textbf{14.58} & \textbf{0.789} & 0.784 & 0.349 & \textbf{0.283} \\
\bottomrule
\end{tabular}
\vspace{-2mm}
\caption{\textbf{Multi-view diffusion.} We compare ToMeSD~\cite{Bolya2023TokenMF} with our token merging method when applied to Zero123++~\cite{shi2023zero123++}.}
\label{tab:mv}
\vspace{-1mm}
\end{table}

In~\cref{tab:dit} and~\cref{fig:pixart}, we show our method significantly outperforms ToMeSD when applied to a diffusion transformer, achieving an FID improvement of \textbf{17–48\%}. This highlights the generalizability of our approach.

In~\cref{tab:mv}, we compare ToMeSD and our token merging method in the context of multi-view diffusion. Our importance-based token merging method consistently shows improved performance over ToMeSD, especially at higher merging ratios. At $r = 0.75$, our method achieves a significant improvement in PSNR (14.58 vs. 13.12) and a much lower LPIPS (0.283 vs. 0.349), highlighting its ability to maintain high output quality even under aggressive token compression. Qualitative examples in~\cref{fig:mv} further validate this, showcasing finer geometrical and textual details in the objects generated by our method.

\begin{figure*}[t]
\begin{center}
\includegraphics[width=1\textwidth]{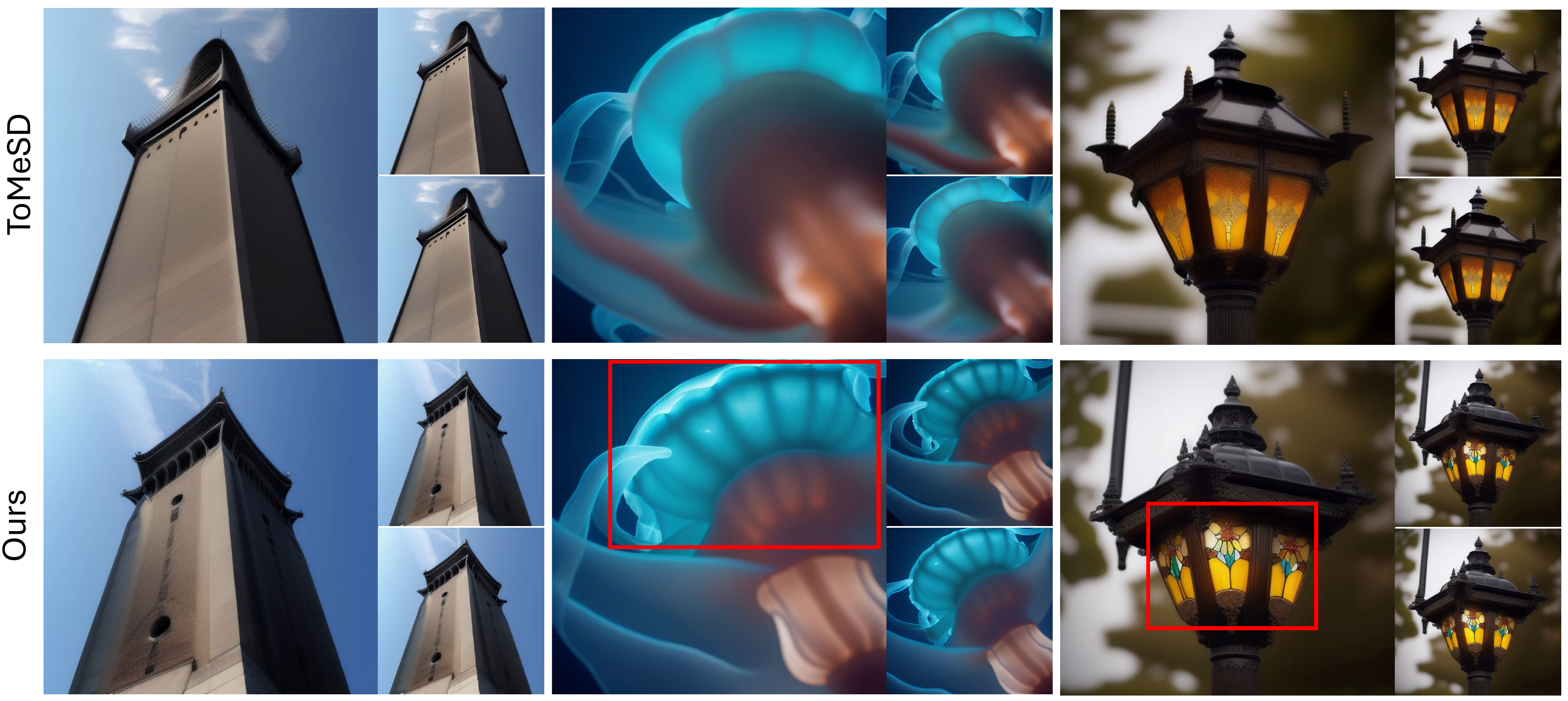}
\end{center}
\vspace{-6mm}
\caption{\textbf{Qualitative comparison of video diffusion.} We apply ToMeSD and our token merging method to the video diffusion model. For each generated video, we show three frames: the first on the left, the 8th at the top right, and the last 16th frame at the bottom right. We use AnimateDiff~\cite{guo2023animatediff} as the base model and a merging ratio of 0.7. Best viewed with zoom-in. Please refer to supplementary for prompts.}
\label{fig:video}
\vspace{-4mm}
\end{figure*}

A similar trend can be observed in~\cref{tab:video} and~\cref{fig:video}, where we extend the comparison to video diffusion. Across these tests, our method consistently performs better than ToMeSD, both in numerical metrics and in visual quality, particularly in preserving object details. 
We observed that merging tokens in the temporal layers significantly reduces the generated dynamics. Nonetheless, we present results with token merging applied for both spatial and temporal layers in~\cref{tab:video2}.

\begin{table}[t]
\centering
\begin{tabular}{ccccccc}
\toprule
 \multirow{2}{*}{$r$} & \multicolumn{2}{c}{Semantic $\uparrow$} & \multicolumn{2}{c}{Quality $\uparrow$} & \multicolumn{2}{c}{Total $\uparrow$} \\
\cmidrule(lr){2-3} \cmidrule(lr){4-5} \cmidrule(lr){6-7}
  & ToMe. & Ours & ToMe. & Ours & ToMe. & Ours \\
\midrule
0.40 & 75.40 & 75.40 & 81.69 & 81.69 & 80.44 & 80.44 \\
0.60 & 74.03 & \textbf{74.51} & 81.58 & \textbf{81.75} & 80.07 & \textbf{80.30} \\
0.70 & 72.03 & \textbf{73.23} & \textbf{81.52} & 81.23 & 79.62 & \textbf{79.63} \\
0.75 & 69.67 & \textbf{71.58} & 80.82 & \textbf{81.00} & 78.59 & \textbf{79.12} \\
\bottomrule
\end{tabular}
\vspace{-2mm}
\caption{\textbf{Video diffusion (spatial).} Token merging applies on only spatial attention layers of AnimateDiff~\cite{guo2023animatediff} with various token merging ratios $r$. We compare our method and ToMeSD~\cite{Bolya2023TokenMF} with VBench scores~\cite{huang2024vbench}.}
\label{tab:video}
\vspace{-1mm}
\end{table}

\begin{table}[t]
\centering
\begin{tabular}{cccc}
\toprule
& Semantic $\uparrow$ & Quality $\uparrow$ & Total $\uparrow$ \\
\midrule
ToMeSD & 72.71 & 81.35 &  79.62 \\
Ours & \textbf{73.52} & \textbf{81.69} &  \textbf{80.06} \\
\bottomrule
\end{tabular}
\vspace{-2mm}
\caption{\textbf{Video diffusion (spatial and temporal)}. Token merging applies on both spatial and temporal attention layers of AnimateDiff~\cite{guo2023animatediff} with a spatial merging ratio of 0.7 and a temporal merging ratio of 0.2. We report VBench scores~\cite{huang2024vbench} of our method in comparison to ToMeSD~\cite{Bolya2023TokenMF}.}
\label{tab:video2}
\vspace{-1mm}
\end{table}

\paragraph{Inference Costs.} We compare the inference costs of our method and ToMeSD when applied to Stable Diffusion 2 in~\cref{tab:cost}. As can be seen, both methods show similar improvements in inference times, GPU usage, and TFLOPs. This demonstrates that our token merging strategy can enhance performance without incurring additional costs.

\begin{table}[t]
\centering
\begin{tabular}{ccccc}
\toprule
\multirow{2}{*}{$r$} & \multicolumn{2}{c}{Latency (s)} & \multirow{2}{*}{Memory (GB)} & \multirow{2}{*}{TFLOPs} \\
\cmidrule(lr){2-3}
& ToMeSD & Ours &  &  \\
\midrule
0   & \multicolumn{2}{c}{8.5 (5.3)} & 7.63 (3.16)	&	4.30 \\
0.3 & 8.0 (5.0) & 8.0 (5.0) & 5.20 (3.16) & 3.83 \\
0.5 & 6.0 (4.7) & 6.1 (4.8)	& 4.05 (3.16) & 3.55 \\
0.7 & 5.7 (4.5) & 5.8 (4.5)	& 3.55 (3.16) & 3.36 \\
\bottomrule
\end{tabular}
\vspace{-2mm}
\caption{Comparison of inference costs when applying our token merging method and ToMeSD~\cite{Bolya2023TokenMF} to Stable Diffusion 2~\cite{rombach2022high}. Numbers in parentheses indicate measurements when memory-efficient attention~\cite{xFormers2022} is enabled.}
\label{tab:cost}
\vspace{-1mm}
\end{table}

\subsection{Ablation Study}
\label{ssec:ablation}

We conduct ablation studies on the text-to-image generation task to examine alternative design choices in our method. As shown in~\cref{tab:abl}, simply using top-k important tokens as destination tokens leads to worse results. Furthermore, not choosing independent tokens exclusively from the important set (w/ global \textit{ind.}), leads to a performance drop.

\begin{table}[t]
\centering
\begin{tabular}{c|cccc}
\toprule
$r$ &  & Ours & w/ top-k \textit{dst} & w/ global \textit{ind.} \\
\midrule
\multirow{2}{*}{0.3} & FID $\downarrow$ & \textbf{12.20} & 12.56   & 12.22 \\
 & CLIP $\uparrow$ & 31.82 & 31.82  & 31.82  \\
 \cmidrule(lr){1-5}
\multirow{2}{*}{0.7} & FID $\downarrow$ & \textbf{16.22}  & 16.29  & 16.43  \\
 & CLIP $\uparrow$ & 31.79  & 31.79 & 31.80  \\
\bottomrule
\end{tabular}
\vspace{-2mm}
\caption{Ablation studies of our method for text-to-image generation. 'w/ top-k \textit{dst}' means top-k rather than random selection from the important token pool for destination tokens. 'w/ global \textit{ind.}' means independent tokens may also be outside the important token pool, instead of solely within it.}
\label{tab:abl}
\vspace{-2mm}
\end{table}

\begin{table}[t]
\centering
\begin{tabular}{ccccccc}
\toprule
 \multirow{3}{*}{$r$} & \multicolumn{3}{c}{FID $\downarrow$} & \multicolumn{3}{c}{CLIP $\uparrow$} \\
\cmidrule(lr){2-4} \cmidrule(lr){5-7}
& \multirow{2}{*}{ToMe} & Ours & Ours & \multirow{2}{*}{ToMe} & Ours & Ours \\
&   & \textit{(CA)} & \textit{(CFG)} &   & \textit{(CA)} & \textit{(CFG)} \\
\midrule
0.30 & 12.20 & \textbf{12.17} & 12.20 & 31.82 & 31.82 & 31.82 \\
0.50 & 13.50 & \textbf{13.38} & 13.42 & 31.79 & 31.82 & \textbf{31.83} \\
0.70 & 17.46 & 16.79 & \textbf{16.22} & 31.78 & 31.79 & \textbf{31.79} \\
0.75 & 20.89 & 18.17 & \textbf{17.75}  & 31.71 & 31.75 & \textbf{31.76} \\
\bottomrule
\end{tabular}
\vspace{-2mm}
\caption{Comparison of token merging methods applied to Stable Diffusion~\cite{rombach2022high}. CA and CFG denote cross-attention and classifier-free guidance as importance signals, respectively.}
\label{tab:ca_sd2}
\vspace{-3mm}
\end{table}

\begin{table}[t]
\centering
\begin{tabular}{c|ccccc}
\toprule
$r$ & $p$ & 0 & 0.2 & 0.4 & 0.8 \\
\midrule
\multirow{2}{*}{0.3} & FID $\downarrow$ & 12.87 & 12.32 & \textbf{12.19} & \textbf{12.19} \\
 & CLIP $\uparrow$ & 31.83 & 31.83 & 31.82 & 31.82 \\
 \cmidrule(lr){1-6}
\multirow{2}{*}{0.7} & FID $\downarrow$ & 16.52 & 16.23 & \textbf{16.22} & 16.42 \\
 & CLIP $\uparrow$ & 31.78 & 31.78 & 31.79 & 31.79 \\
\bottomrule
\end{tabular}
\vspace{-2mm}
\caption{\textbf{Choice of $p$.} We show the results of our method for the text-to-image generation task with different values of $p$, which determines the important token pool size.}
\label{tab:ablation_p}
\end{table}

\paragraph{Cross-Attention Maps for Token Importance.} In principle, our method can be used with any per-token importance scores. As shown in~\cref{tab:ca_sd2}, we use our method with cross-attention maps instead of CFG. While this approach may require more memory compared to CFG, it remains a strong alternative and highlights the generalizability of our method.

\paragraph{Choice of $p$.} Our important token pool size is $(1 - r)\cdot(1 + p)$, where $r$ is the token merging ratio. \cref{tab:ablation_p} shows that our method remains robust to the choice of $p$, provided that $p$ is within a reasonable range.
\section{Conclusions}
\label{sec:conlusions}

We propose an importance-based token merging method for generation tasks, which maintains generation quality while reducing inference latency. 
We utilize token importance to strategically allocate computational resources to regions of high relevance to the input condition, thereby enhancing the fidelity of the generated outputs.
This novel, simple, and intuitive strategy accelerates various models for free with no modifications needed. Notably, we identify classifier-free guidance as an effective token importance indicator.
Our method achieves state-of-the-art performance across diverse tasks, including text-to-image synthesis, multi-view generation, and video generation, highlighting its effectiveness and versatility. 

\clearpage
{
    \small
    \bibliographystyle{ieeenat_fullname}
    \bibliography{main}

\begin{thebibliography}{119}
\providecommand{\natexlab}[1]{#1}
\providecommand{\url}[1]{\texttt{#1}}
\expandafter\ifx\csname urlstyle\endcsname\relax
  \providecommand{\doi}[1]{doi: #1}\else
  \providecommand{\doi}{doi: \begingroup \urlstyle{rm}\Url}\fi

\bibitem[Balaji et~al.(2022)Balaji, Nah, Huang, Vahdat, Song, Zhang, Kreis, Aittala, Aila, Laine, et~al.]{balaji2022ediff}
Yogesh Balaji, Seungjun Nah, Xun Huang, Arash Vahdat, Jiaming Song, Qinsheng Zhang, Karsten Kreis, Miika Aittala, Timo Aila, Samuli Laine, et~al.
\newblock ediff-i: Text-to-image diffusion models with an ensemble of expert denoisers.
\newblock \emph{arXiv preprint arXiv:2211.01324}, 2022.

\bibitem[Blattmann et~al.(2023)Blattmann, Dockhorn, Kulal, Mendelevitch, Kilian, Lorenz, Levi, English, Voleti, Letts, et~al.]{blattmann2023stable}
Andreas Blattmann, Tim Dockhorn, Sumith Kulal, Daniel Mendelevitch, Maciej Kilian, Dominik Lorenz, Yam Levi, Zion English, Vikram Voleti, Adam Letts, et~al.
\newblock Stable video diffusion: Scaling latent video diffusion models to large datasets.
\newblock \emph{arXiv preprint arXiv:2311.15127}, 2023.

\bibitem[Bolya and Hoffman(2023)]{Bolya2023TokenMF}
Daniel Bolya and Judy Hoffman.
\newblock Token merging for fast stable diffusion.
\newblock \emph{2023 IEEE/CVF Conference on Computer Vision and Pattern Recognition Workshops (CVPRW)}, pages 4599--4603, 2023.

\bibitem[Bolya et~al.(2022)Bolya, Fu, Dai, Zhang, Feichtenhofer, and Hoffman]{Bolya2022TokenMY}
Daniel Bolya, Cheng-Yang Fu, Xiaoliang Dai, Peizhao Zhang, Christoph Feichtenhofer, and Judy Hoffman.
\newblock Token merging: Your vit but faster.
\newblock \emph{ArXiv}, abs/2210.09461, 2022.

\bibitem[Brooks et~al.(2023)Brooks, Holynski, and Efros]{brooks2023instructpix2pix}
Tim Brooks, Aleksander Holynski, and Alexei~A Efros.
\newblock Instructpix2pix: Learning to follow image editing instructions.
\newblock In \emph{Proceedings of the IEEE/CVF conference on computer vision and pattern recognition}, pages 18392--18402, 2023.

\bibitem[Brooks et~al.(2024)Brooks, Peebles, Holmes, DePue, Guo, Jing, Schnurr, Taylor, Luhman, Luhman, et~al.]{brooks2024video}
Tim Brooks, Bill Peebles, Connor Holmes, Will DePue, Yufei Guo, Li Jing, David Schnurr, Joe Taylor, Troy Luhman, Eric Luhman, et~al.
\newblock Video generation models as world simulators. 2024.
\newblock \emph{URL https://openai. com/research/video-generation-models-as-world-simulators}, 3, 2024.

\bibitem[Chen et~al.(2023)Chen, Yu, Ge, Yao, Xie, Wu, Wang, Kwok, Luo, Lu, and Li]{chen2023pixartalpha}
Junsong Chen, Jincheng Yu, Chongjian Ge, Lewei Yao, Enze Xie, Yue Wu, Zhongdao Wang, James Kwok, Ping Luo, Huchuan Lu, and Zhenguo Li.
\newblock Pixart-$\alpha$: Fast training of diffusion transformer for photorealistic text-to-image synthesis, 2023.

\bibitem[Chen et~al.(2024{\natexlab{a}})Chen, Meng, Tang, Ma, Jiang, Wang, Wang, and Zhu]{chen2024q}
Lei Chen, Yuan Meng, Chen Tang, Xinzhu Ma, Jingyan Jiang, Xin Wang, Zhi Wang, and Wenwu Zhu.
\newblock Q-dit: Accurate post-training quantization for diffusion transformers.
\newblock \emph{arXiv preprint arXiv:2406.17343}, 2024{\natexlab{a}}.

\bibitem[Chen et~al.(2024{\natexlab{b}})Chen, Shen, Ye, Cao, Tu, Bouganis, Zhao, and Chen]{chen2024delta}
Pengtao Chen, Mingzhu Shen, Peng Ye, Jianjian Cao, Chongjun Tu, Christos-Savvas Bouganis, Yiren Zhao, and Tao Chen.
\newblock Delta-dit: A training-free acceleration method tailored for diffusion transformers.
\newblock \emph{arXiv preprint arXiv:2406.01125}, 2024{\natexlab{b}}.

\bibitem[Chen et~al.(2021)Chen, Cheng, Gan, Yuan, Zhang, and Wang]{chen2021chasing}
Tianlong Chen, Yu Cheng, Zhe Gan, Lu Yuan, Lei Zhang, and Zhangyang Wang.
\newblock Chasing sparsity in vision transformers: An end-to-end exploration.
\newblock \emph{Advances in Neural Information Processing Systems}, 34:\penalty0 19974--19988, 2021.

\bibitem[Choi et~al.(2024)Choi, Lee, Chu, Choi, and Kim]{choi2024vid}
Joonmyung Choi, Sanghyeok Lee, Jaewon Chu, Minhyuk Choi, and Hyunwoo~J Kim.
\newblock vid-tldr: Training free token merging for light-weight video transformer.
\newblock In \emph{Proceedings of the IEEE/CVF Conference on Computer Vision and Pattern Recognition}, pages 18771--18781, 2024.

\bibitem[Deng et~al.(2024)Deng, Li, Wang, Gu, Xu, and Huang]{deng2024vq4dit}
Juncan Deng, Shuaiting Li, Zeyu Wang, Hong Gu, Kedong Xu, and Kejie Huang.
\newblock Vq4dit: Efficient post-training vector quantization for diffusion transformers.
\newblock \emph{arXiv preprint arXiv:2408.17131}, 2024.

\bibitem[Dhariwal and Nichol(2021)]{dhariwal2021diffusion}
Prafulla Dhariwal and Alexander Nichol.
\newblock Diffusion models beat gans on image synthesis.
\newblock \emph{Advances in neural information processing systems}, 34:\penalty0 8780--8794, 2021.

\bibitem[Downs et~al.(2022)Downs, Francis, Koenig, Kinman, Hickman, Reymann, McHugh, and Vanhoucke]{downs2022google}
Laura Downs, Anthony Francis, Nate Koenig, Brandon Kinman, Ryan Hickman, Krista Reymann, Thomas~B McHugh, and Vincent Vanhoucke.
\newblock Google scanned objects: A high-quality dataset of 3d scanned household items.
\newblock In \emph{2022 International Conference on Robotics and Automation (ICRA)}, pages 2553--2560. IEEE, 2022.

\bibitem[Fang et~al.(2023)Fang, Ma, and Wang]{fang2023structural}
Gongfan Fang, Xinyin Ma, and Xinchao Wang.
\newblock Structural pruning for diffusion models.
\newblock In \emph{Advances in Neural Information Processing Systems}, 2023.

\bibitem[Fayyaz et~al.(2022)Fayyaz, Koohpayegani, Jafari, Sengupta, Joze, Sommerlade, Pirsiavash, and Gall]{fayyaz2022adaptive}
Mohsen Fayyaz, Soroush~Abbasi Koohpayegani, Farnoush~Rezaei Jafari, Sunando Sengupta, Hamid Reza~Vaezi Joze, Eric Sommerlade, Hamed Pirsiavash, and J{\"u}rgen Gall.
\newblock Adaptive token sampling for efficient vision transformers.
\newblock In \emph{European Conference on Computer Vision}, pages 396--414. Springer, 2022.

\bibitem[Graikos et~al.(2024)Graikos, Yellapragada, Le, Kapse, Prasanna, Saltz, and Samaras]{graikos2024learned}
Alexandros Graikos, Srikar Yellapragada, Minh-Quan Le, Saarthak Kapse, Prateek Prasanna, Joel Saltz, and Dimitris Samaras.
\newblock Learned representation-guided diffusion models for large-image generation.
\newblock In \emph{Proceedings of the IEEE/CVF Conference on Computer Vision and Pattern Recognition}, pages 8532--8542, 2024.

\bibitem[Guo et~al.(2023)Guo, Yang, Rao, Liang, Wang, Qiao, Agrawala, Lin, and Dai]{guo2023animatediff}
Yuwei Guo, Ceyuan Yang, Anyi Rao, Zhengyang Liang, Yaohui Wang, Yu Qiao, Maneesh Agrawala, Dahua Lin, and Bo Dai.
\newblock Animatediff: Animate your personalized text-to-image diffusion models without specific tuning.
\newblock \emph{arXiv preprint arXiv:2307.04725}, 2023.

\bibitem[Habibian et~al.(2024)Habibian, Ghodrati, Fathima, Sautiere, Garrepalli, Porikli, and Petersen]{habibian2024clockwork}
Amirhossein Habibian, Amir Ghodrati, Noor Fathima, Guillaume Sautiere, Risheek Garrepalli, Fatih Porikli, and Jens Petersen.
\newblock Clockwork diffusion: Efficient generation with model-step distillation.
\newblock In \emph{Proceedings of the IEEE/CVF Conference on Computer Vision and Pattern Recognition}, pages 8352--8361, 2024.

\bibitem[Haurum et~al.(2022)Haurum, Madadi, Escalera, and Moeslund]{haurum2022multi}
Joakim~Bruslund Haurum, Meysam Madadi, Sergio Escalera, and Thomas~B Moeslund.
\newblock Multi-scale hybrid vision transformer and sinkhorn tokenizer for sewer defect classification.
\newblock \emph{Automation in Construction}, 144:\penalty0 104614, 2022.

\bibitem[Haurum et~al.(2023)Haurum, Escalera, Taylor, and Moeslund]{haurum2023tokens}
Joakim~Bruslund Haurum, Sergio Escalera, Graham~W Taylor, and Thomas~B Moeslund.
\newblock Which tokens to use? investigating token reduction in vision transformers.
\newblock In \emph{Proceedings of the IEEE/CVF International Conference on Computer Vision}, pages 773--783, 2023.

\bibitem[Haurum et~al.(2025)Haurum, Escalera, Taylor, and Moeslund]{haurum2025agglomerative}
Joakim~Bruslund Haurum, Sergio Escalera, Graham~W Taylor, and Thomas~B Moeslund.
\newblock Agglomerative token clustering.
\newblock In \emph{European Conference on Computer Vision}, pages 200--218. Springer, 2025.

\bibitem[He et~al.(2024)He, Liu, Liu, Wu, Zhou, and Zhuang]{he2024ptqd}
Yefei He, Luping Liu, Jing Liu, Weijia Wu, Hong Zhou, and Bohan Zhuang.
\newblock Ptqd: Accurate post-training quantization for diffusion models.
\newblock \emph{Advances in Neural Information Processing Systems}, 36, 2024.

\bibitem[Hessel et~al.(2021)Hessel, Holtzman, Forbes, Bras, and Choi]{hessel2021clipscore}
Jack Hessel, Ari Holtzman, Maxwell Forbes, Ronan~Le Bras, and Yejin Choi.
\newblock Clipscore: A reference-free evaluation metric for image captioning.
\newblock \emph{arXiv preprint arXiv:2104.08718}, 2021.

\bibitem[Heusel et~al.(2017)Heusel, Ramsauer, Unterthiner, Nessler, and Hochreiter]{heusel2017gans}
Martin Heusel, Hubert Ramsauer, Thomas Unterthiner, Bernhard Nessler, and Sepp Hochreiter.
\newblock Gans trained by a two time-scale update rule converge to a local nash equilibrium.
\newblock \emph{Advances in neural information processing systems}, 30, 2017.

\bibitem[Ho and Salimans(2022)]{ho2022classifier}
Jonathan Ho and Tim Salimans.
\newblock Classifier-free diffusion guidance.
\newblock \emph{arXiv preprint arXiv:2207.12598}, 2022.

\bibitem[Ho et~al.(2020)Ho, Jain, and Abbeel]{ho2020denoising}
Jonathan Ho, Ajay Jain, and Pieter Abbeel.
\newblock Denoising diffusion probabilistic models.
\newblock \emph{Advances in neural information processing systems}, 33:\penalty0 6840--6851, 2020.

\bibitem[Huang et~al.(2023)Huang, Huang, Yang, Ren, Liu, Li, Ye, Liu, Yin, and Zhao]{huang2023make}
Rongjie Huang, Jiawei Huang, Dongchao Yang, Yi Ren, Luping Liu, Mingze Li, Zhenhui Ye, Jinglin Liu, Xiang Yin, and Zhou Zhao.
\newblock Make-an-audio: Text-to-audio generation with prompt-enhanced diffusion models.
\newblock In \emph{International Conference on Machine Learning}, pages 13916--13932. PMLR, 2023.

\bibitem[Huang et~al.(2024)Huang, He, Yu, Zhang, Si, Jiang, Zhang, Wu, Jin, Chanpaisit, et~al.]{huang2024vbench}
Ziqi Huang, Yinan He, Jiashuo Yu, Fan Zhang, Chenyang Si, Yuming Jiang, Yuanhan Zhang, Tianxing Wu, Qingyang Jin, Nattapol Chanpaisit, et~al.
\newblock Vbench: Comprehensive benchmark suite for video generative models.
\newblock In \emph{Proceedings of the IEEE/CVF Conference on Computer Vision and Pattern Recognition}, pages 21807--21818, 2024.

\bibitem[Jang et~al.(2016)Jang, Gu, and Poole]{jang2016categorical}
Eric Jang, Shixiang Gu, and Ben Poole.
\newblock Categorical reparameterization with gumbel-softmax.
\newblock \emph{arXiv preprint arXiv:1611.01144}, 2016.

\bibitem[Jin et~al.(2024)Jin, Takanobu, Zhang, Cao, and Yuan]{jin2024chat}
Peng Jin, Ryuichi Takanobu, Wancai Zhang, Xiaochun Cao, and Li Yuan.
\newblock Chat-univi: Unified visual representation empowers large language models with image and video understanding.
\newblock In \emph{Proceedings of the IEEE/CVF Conference on Computer Vision and Pattern Recognition}, pages 13700--13710, 2024.

\bibitem[Kahatapitiya et~al.(2024)Kahatapitiya, Liu, He, Liu, Jia, Ryoo, and Xie]{kahatapitiya2024adaptive}
Kumara Kahatapitiya, Haozhe Liu, Sen He, Ding Liu, Menglin Jia, Michael~S Ryoo, and Tian Xie.
\newblock Adaptive caching for faster video generation with diffusion transformers.
\newblock \emph{arXiv preprint arXiv:2411.02397}, 2024.

\bibitem[Kahatapitiya et~al.(2025)Kahatapitiya, Karjauv, Abati, Porikli, Asano, and Habibian]{kahatapitiya2025object}
Kumara Kahatapitiya, Adil Karjauv, Davide Abati, Fatih Porikli, Yuki~M Asano, and Amirhossein Habibian.
\newblock Object-centric diffusion for efficient video editing.
\newblock In \emph{European Conference on Computer Vision}, pages 91--108. Springer, 2025.

\bibitem[Karras et~al.(2025)Karras, Aittala, Kynk{\"a}{\"a}nniemi, Lehtinen, Aila, and Laine]{karras2025guiding}
Tero Karras, Miika Aittala, Tuomas Kynk{\"a}{\"a}nniemi, Jaakko Lehtinen, Timo Aila, and Samuli Laine.
\newblock Guiding a diffusion model with a bad version of itself.
\newblock \emph{Advances in Neural Information Processing Systems}, 37:\penalty0 52996--53021, 2025.

\bibitem[Kienzle et~al.(2024)Kienzle, Kantonis, Sch{\"o}n, and Lienhart]{kienzle2024segformer++}
Daniel Kienzle, Marco Kantonis, Robin Sch{\"o}n, and Rainer Lienhart.
\newblock Segformer++: Efficient token-merging strategies for high-resolution semantic segmentation.
\newblock \emph{arXiv preprint arXiv:2405.14467}, 2024.

\bibitem[Kim et~al.(2023)Kim, Song, Castells, and Choi]{kim2023architectural}
Bo-Kyeong Kim, Hyoung-Kyu Song, Thibault Castells, and Shinkook Choi.
\newblock On architectural compression of text-to-image diffusion models.
\newblock 2023.

\bibitem[Kim et~al.(2024)Kim, Gao, Hsu, Shen, and Jin]{kim2024token}
Minchul Kim, Shangqian Gao, Yen-Chang Hsu, Yilin Shen, and Hongxia Jin.
\newblock Token fusion: Bridging the gap between token pruning and token merging.
\newblock In \emph{Proceedings of the IEEE/CVF Winter Conference on Applications of Computer Vision}, pages 1383--1392, 2024.

\bibitem[Kong et~al.(2020)Kong, Ping, Huang, Zhao, and Catanzaro]{kong2020diffwave}
Zhifeng Kong, Wei Ping, Jiaji Huang, Kexin Zhao, and Bryan Catanzaro.
\newblock Diffwave: A versatile diffusion model for audio synthesis.
\newblock \emph{arXiv preprint arXiv:2009.09761}, 2020.

\bibitem[Kong et~al.(2022)Kong, Dong, Ma, Meng, Niu, Sun, Shen, Yuan, Ren, Tang, et~al.]{kong2022spvit}
Zhenglun Kong, Peiyan Dong, Xiaolong Ma, Xin Meng, Wei Niu, Mengshu Sun, Xuan Shen, Geng Yuan, Bin Ren, Hao Tang, et~al.
\newblock Spvit: Enabling faster vision transformers via latency-aware soft token pruning.
\newblock In \emph{European conference on computer vision}, pages 620--640. Springer, 2022.

\bibitem[Lab and etc.(2024)]{pku_yuan_lab_and_tuzhan_ai_etc_2024_10948109}
PKU-Yuan Lab and Tuzhan~AI etc.
\newblock Open-sora-plan, 2024.

\bibitem[Lee and Hong(2024)]{leelearning}
Dong~Hoon Lee and Seunghoon Hong.
\newblock Learning to merge tokens via decoupled embedding for efficient vision transformers.
\newblock In \emph{The Thirty-eighth Annual Conference on Neural Information Processing Systems}, 2024.

\bibitem[Lefaudeux et~al.(2022)Lefaudeux, Massa, Liskovich, Xiong, Caggiano, Naren, Xu, Hu, Tintore, Zhang, Labatut, Haziza, Wehrstedt, Reizenstein, and Sizov]{xFormers2022}
Benjamin Lefaudeux, Francisco Massa, Diana Liskovich, Wenhan Xiong, Vittorio Caggiano, Sean Naren, Min Xu, Jieru Hu, Marta Tintore, Susan Zhang, Patrick Labatut, Daniel Haziza, Luca Wehrstedt, Jeremy Reizenstein, and Grigory Sizov.
\newblock xformers: A modular and hackable transformer modelling library.
\newblock \url{https://github.com/facebookresearch/xformers}, 2022.

\bibitem[Li et~al.(2022{\natexlab{a}})Li, Zheng, Wang, Li, Zheng, and Tao]{li20223ddesigner}
Gang Li, Heliang Zheng, Chaoyue Wang, Chang Li, Changwen Zheng, and Dacheng Tao.
\newblock 3ddesigner: Towards photorealistic 3d object generation and editing with text-guided diffusion models.
\newblock \emph{arXiv preprint arXiv:2211.14108}, 2022{\natexlab{a}}.

\bibitem[Li et~al.(2022{\natexlab{b}})Li, Thorsley, and Hassoun]{li2022sait}
Ling Li, David Thorsley, and Joseph Hassoun.
\newblock Sait: Sparse vision transformers through adaptive token pruning.
\newblock \emph{arXiv preprint arXiv:2210.05832}, 2022{\natexlab{b}}.

\bibitem[Li et~al.(2023{\natexlab{a}})Li, Hu, Khan, Li, Yang, Wang, Cheng, and Yang]{li2023faster}
Senmao Li, Taihang Hu, Fahad~Shahbaz Khan, Linxuan Li, Shiqi Yang, Yaxing Wang, Ming-Ming Cheng, and Jian Yang.
\newblock Faster diffusion: Rethinking the role of unet encoder in diffusion models.
\newblock \emph{CoRR}, 2023{\natexlab{a}}.

\bibitem[Li et~al.(2024{\natexlab{a}})Li, Yuan, Liu, Tang, Wang, Qin, Zhu, and Zhang]{li2024tokenpacker}
Wentong Li, Yuqian Yuan, Jian Liu, Dongqi Tang, Song Wang, Jie Qin, Jianke Zhu, and Lei Zhang.
\newblock Tokenpacker: Efficient visual projector for multimodal llm.
\newblock \emph{arXiv preprint arXiv:2407.02392}, 2024{\natexlab{a}}.

\bibitem[Li et~al.(2023{\natexlab{b}})Li, Liu, Lian, Yang, Dong, Kang, Zhang, and Keutzer]{li2023q}
Xiuyu Li, Yijiang Liu, Long Lian, Huanrui Yang, Zhen Dong, Daniel Kang, Shanghang Zhang, and Kurt Keutzer.
\newblock Q-diffusion: Quantizing diffusion models.
\newblock In \emph{Proceedings of the IEEE/CVF International Conference on Computer Vision}, pages 17535--17545, 2023{\natexlab{b}}.

\bibitem[Li et~al.(2024{\natexlab{b}})Li, Ma, Yang, and Yang]{li2024vidtome}
Xirui Li, Chao Ma, Xiaokang Yang, and Ming-Hsuan Yang.
\newblock Vidtome: Video token merging for zero-shot video editing.
\newblock In \emph{Proceedings of the IEEE/CVF Conference on Computer Vision and Pattern Recognition}, pages 7486--7495, 2024{\natexlab{b}}.

\bibitem[Li et~al.(2024{\natexlab{c}})Li, Wang, Jin, Hu, Chemerys, Fu, Wang, Tulyakov, and Ren]{li2024snapfusion}
Yanyu Li, Huan Wang, Qing Jin, Ju Hu, Pavlo Chemerys, Yun Fu, Yanzhi Wang, Sergey Tulyakov, and Jian Ren.
\newblock Snapfusion: Text-to-image diffusion model on mobile devices within two seconds.
\newblock \emph{Advances in Neural Information Processing Systems}, 36, 2024{\natexlab{c}}.

\bibitem[Liang et~al.(2024)Liang, Kodaira, Xu, Tomizuka, Keutzer, and Marculescu]{liang2024looking}
Feng Liang, Akio Kodaira, Chenfeng Xu, Masayoshi Tomizuka, Kurt Keutzer, and Diana Marculescu.
\newblock Looking backward: Streaming video-to-video translation with feature banks.
\newblock \emph{arXiv preprint arXiv:2405.15757}, 2024.

\bibitem[Liang et~al.(2022)Liang, Ge, Tong, Song, Wang, and Xie]{liang2022not}
Youwei Liang, Chongjian Ge, Zhan Tong, Yibing Song, Jue Wang, and Pengtao Xie.
\newblock Not all patches are what you need: Expediting vision transformers via token reorganizations.
\newblock \emph{arXiv preprint arXiv:2202.07800}, 2022.

\bibitem[Lin et~al.(2014)Lin, Maire, Belongie, Hays, Perona, Ramanan, Doll{\'a}r, and Zitnick]{lin2014microsoft}
Tsung-Yi Lin, Michael Maire, Serge Belongie, James Hays, Pietro Perona, Deva Ramanan, Piotr Doll{\'a}r, and C~Lawrence Zitnick.
\newblock Microsoft coco: Common objects in context.
\newblock In \emph{Computer Vision--ECCV 2014: 13th European Conference, Zurich, Switzerland, September 6-12, 2014, Proceedings, Part V 13}, pages 740--755. Springer, 2014.

\bibitem[Liu et~al.(2022)Liu, Ren, Lin, and Zhao]{liu2022pseudo}
Luping Liu, Yi Ren, Zhijie Lin, and Zhou Zhao.
\newblock Pseudo numerical methods for diffusion models on manifolds.
\newblock \emph{arXiv preprint arXiv:2202.09778}, 2022.

\bibitem[Liu et~al.(2023{\natexlab{a}})Liu, Wu, Van~Hoorick, Tokmakov, Zakharov, and Vondrick]{liu2023zero}
Ruoshi Liu, Rundi Wu, Basile Van~Hoorick, Pavel Tokmakov, Sergey Zakharov, and Carl Vondrick.
\newblock Zero-1-to-3: Zero-shot one image to 3d object.
\newblock In \emph{Proceedings of the IEEE/CVF international conference on computer vision}, pages 9298--9309, 2023{\natexlab{a}}.

\bibitem[Liu et~al.(2023{\natexlab{b}})Liu, Zhang, Ma, Peng, et~al.]{liu2023instaflow}
Xingchao Liu, Xiwen Zhang, Jianzhu Ma, Jian Peng, et~al.
\newblock Instaflow: One step is enough for high-quality diffusion-based text-to-image generation.
\newblock In \emph{The Twelfth International Conference on Learning Representations}, 2023{\natexlab{b}}.

\bibitem[Liu et~al.(2023{\natexlab{c}})Liu, Lin, Zeng, Long, Liu, Komura, and Wang]{liu2023syncdreamer}
Yuan Liu, Cheng Lin, Zijiao Zeng, Xiaoxiao Long, Lingjie Liu, Taku Komura, and Wenping Wang.
\newblock Syncdreamer: Generating multiview-consistent images from a single-view image.
\newblock \emph{arXiv preprint arXiv:2309.03453}, 2023{\natexlab{c}}.

\bibitem[Liu et~al.(2024)Liu, Gehrig, Messikommer, Cannici, and Scaramuzza]{liu2024revisiting}
Yifei Liu, Mathias Gehrig, Nico Messikommer, Marco Cannici, and Davide Scaramuzza.
\newblock Revisiting token pruning for object detection and instance segmentation.
\newblock In \emph{Proceedings of the IEEE/CVF Winter Conference on Applications of Computer Vision}, pages 2658--2668, 2024.

\bibitem[Long et~al.(2023)Long, Zhao, Pi, Wang, and Wang]{long2023beyond}
Sifan Long, Zhen Zhao, Jimin Pi, Shengsheng Wang, and Jingdong Wang.
\newblock Beyond attentive tokens: Incorporating token importance and diversity for efficient vision transformers.
\newblock In \emph{Proceedings of the IEEE/CVF Conference on Computer Vision and Pattern Recognition}, pages 10334--10343, 2023.

\bibitem[Long et~al.(2024)Long, Guo, Lin, Liu, Dou, Liu, Ma, Zhang, Habermann, Theobalt, et~al.]{long2024wonder3d}
Xiaoxiao Long, Yuan-Chen Guo, Cheng Lin, Yuan Liu, Zhiyang Dou, Lingjie Liu, Yuexin Ma, Song-Hai Zhang, Marc Habermann, Christian Theobalt, et~al.
\newblock Wonder3d: Single image to 3d using cross-domain diffusion.
\newblock In \emph{Proceedings of the IEEE/CVF Conference on Computer Vision and Pattern Recognition}, pages 9970--9980, 2024.

\bibitem[Lu et~al.(2022{\natexlab{a}})Lu, Zhou, Bao, Chen, Li, and Zhu]{lu2022dpm}
Cheng Lu, Yuhao Zhou, Fan Bao, Jianfei Chen, Chongxuan Li, and Jun Zhu.
\newblock Dpm-solver: A fast ode solver for diffusion probabilistic model sampling in around 10 steps.
\newblock \emph{Advances in Neural Information Processing Systems}, 35:\penalty0 5775--5787, 2022{\natexlab{a}}.

\bibitem[Lu et~al.(2022{\natexlab{b}})Lu, Zhou, Bao, Chen, Li, and Zhu]{lu2022dpm_pp}
Cheng Lu, Yuhao Zhou, Fan Bao, Jianfei Chen, Chongxuan Li, and Jun Zhu.
\newblock Dpm-solver++: Fast solver for guided sampling of diffusion probabilistic models.
\newblock \emph{arXiv preprint arXiv:2211.01095}, 2022{\natexlab{b}}.

\bibitem[Lu et~al.(2024)Lu, Zheng, Xia, and Wang]{lutoma}
Wenbo Lu, Shaoyi Zheng, Yuxuan Xia, and Shengjie Wang.
\newblock Toma: Token merging with attention for diffusion models.
\newblock 2024.

\bibitem[Luo and Hu(2021)]{luo2021diffusion}
Shitong Luo and Wei Hu.
\newblock Diffusion probabilistic models for 3d point cloud generation.
\newblock In \emph{Proceedings of the IEEE/CVF conference on computer vision and pattern recognition}, pages 2837--2845, 2021.

\bibitem[Luo et~al.(2023{\natexlab{a}})Luo, Tan, Huang, Li, and Zhao]{luo2023latent}
Simian Luo, Yiqin Tan, Longbo Huang, Jian Li, and Hang Zhao.
\newblock Latent consistency models: Synthesizing high-resolution images with few-step inference.
\newblock \emph{arXiv preprint arXiv:2310.04378}, 2023{\natexlab{a}}.

\bibitem[Luo et~al.(2023{\natexlab{b}})Luo, Tan, Patil, Gu, von Platen, Passos, Huang, Li, and Zhao]{luo2023lcm}
Simian Luo, Yiqin Tan, Suraj Patil, Daniel Gu, Patrick von Platen, Apolin{\'a}rio Passos, Longbo Huang, Jian Li, and Hang Zhao.
\newblock Lcm-lora: A universal stable-diffusion acceleration module.
\newblock \emph{arXiv preprint arXiv:2311.05556}, 2023{\natexlab{b}}.

\bibitem[Lv et~al.(2024)Lv, Si, Song, Yang, Qiao, Liu, and Wong]{lv2024fastercache}
Zhengyao Lv, Chenyang Si, Junhao Song, Zhenyu Yang, Yu Qiao, Ziwei Liu, and Kwan-Yee~K Wong.
\newblock Fastercache: Training-free video diffusion model acceleration with high quality.
\newblock \emph{arXiv preprint arXiv:2410.19355}, 2024.

\bibitem[Ma et~al.(2024)Ma, Fang, and Wang]{ma2024deepcache}
Xinyin Ma, Gongfan Fang, and Xinchao Wang.
\newblock Deepcache: Accelerating diffusion models for free.
\newblock In \emph{Proceedings of the IEEE/CVF Conference on Computer Vision and Pattern Recognition}, pages 15762--15772, 2024.

\bibitem[Marin et~al.(2023)Marin, Chang, Ranjan, Prabhu, Rastegari, and Tuzel]{marin2023token}
Dmitrii Marin, Jen-Hao~Rick Chang, Anurag Ranjan, Anish Prabhu, Mohammad Rastegari, and Oncel Tuzel.
\newblock Token pooling in vision transformers for image classification.
\newblock In \emph{Proceedings of the IEEE/CVF Winter Conference on Applications of Computer Vision}, pages 12--21, 2023.

\bibitem[Meng et~al.(2021)Meng, He, Song, Song, Wu, Zhu, and Ermon]{meng2021sdedit}
Chenlin Meng, Yutong He, Yang Song, Jiaming Song, Jiajun Wu, Jun-Yan Zhu, and Stefano Ermon.
\newblock Sdedit: Guided image synthesis and editing with stochastic differential equations.
\newblock \emph{arXiv preprint arXiv:2108.01073}, 2021.

\bibitem[Meng et~al.(2023)Meng, Rombach, Gao, Kingma, Ermon, Ho, and Salimans]{meng2023distillation}
Chenlin Meng, Robin Rombach, Ruiqi Gao, Diederik Kingma, Stefano Ermon, Jonathan Ho, and Tim Salimans.
\newblock On distillation of guided diffusion models.
\newblock In \emph{Proceedings of the IEEE/CVF Conference on Computer Vision and Pattern Recognition}, pages 14297--14306, 2023.

\bibitem[Nichol et~al.(2021)Nichol, Dhariwal, Ramesh, Shyam, Mishkin, McGrew, Sutskever, and Chen]{nichol2021glide}
Alex Nichol, Prafulla Dhariwal, Aditya Ramesh, Pranav Shyam, Pamela Mishkin, Bob McGrew, Ilya Sutskever, and Mark Chen.
\newblock Glide: Towards photorealistic image generation and editing with text-guided diffusion models.
\newblock \emph{arXiv preprint arXiv:2112.10741}, 2021.

\bibitem[Nichol et~al.(2022)Nichol, Jun, Dhariwal, Mishkin, and Chen]{nichol2022point}
Alex Nichol, Heewoo Jun, Prafulla Dhariwal, Pamela Mishkin, and Mark Chen.
\newblock Point-e: A system for generating 3d point clouds from complex prompts.
\newblock \emph{arXiv preprint arXiv:2212.08751}, 2022.

\bibitem[Pan et~al.(2023)Pan, Panda, Feris, and Oliva]{pan2023interpretability}
Bowen Pan, Rameswar Panda, Rogerio~Schmidt Feris, and Aude~Jeanne Oliva.
\newblock Interpretability-aware redundancy reduction for vision transformers, 2023.
\newblock US Patent App. 17/559,053.

\bibitem[Parmar et~al.(2022)Parmar, Zhang, and Zhu]{parmar2021cleanfid}
Gaurav Parmar, Richard Zhang, and Jun-Yan Zhu.
\newblock On aliased resizing and surprising subtleties in gan evaluation.
\newblock In \emph{CVPR}, 2022.

\bibitem[Peebles and Xie(2023)]{peebles2023scalable}
William Peebles and Saining Xie.
\newblock Scalable diffusion models with transformers.
\newblock In \emph{Proceedings of the IEEE/CVF International Conference on Computer Vision}, pages 4195--4205, 2023.

\bibitem[Pernias et~al.(2023)Pernias, Rampas, Richter, Pal, and Aubreville]{pernias2023wurstchen}
Pablo Pernias, Dominic Rampas, Mats~L Richter, Christopher~J Pal, and Marc Aubreville.
\newblock W{\"u}rstchen: An efficient architecture for large-scale text-to-image diffusion models.
\newblock \emph{arXiv preprint arXiv:2306.00637}, 2023.

\bibitem[Poole et~al.(2022)Poole, Jain, Barron, and Mildenhall]{poole2022dreamfusion}
Ben Poole, Ajay Jain, Jonathan~T Barron, and Ben Mildenhall.
\newblock Dreamfusion: Text-to-3d using 2d diffusion.
\newblock \emph{arXiv preprint arXiv:2209.14988}, 2022.

\bibitem[Radford et~al.(2021)Radford, Kim, Hallacy, Ramesh, Goh, Agarwal, Sastry, Askell, Mishkin, Clark, et~al.]{radford2021learning}
Alec Radford, Jong~Wook Kim, Chris Hallacy, Aditya Ramesh, Gabriel Goh, Sandhini Agarwal, Girish Sastry, Amanda Askell, Pamela Mishkin, Jack Clark, et~al.
\newblock Learning transferable visual models from natural language supervision.
\newblock In \emph{International conference on machine learning}, pages 8748--8763. PMLR, 2021.

\bibitem[Ramesh et~al.(2022)Ramesh, Dhariwal, Nichol, Chu, and Chen]{Ramesh2022HierarchicalTI}
Aditya Ramesh, Prafulla Dhariwal, Alex Nichol, Casey Chu, and Mark Chen.
\newblock Hierarchical text-conditional image generation with clip latents.
\newblock \emph{ArXiv}, abs/2204.06125, 2022.

\bibitem[Rao et~al.(2021)Rao, Zhao, Liu, Lu, Zhou, and Hsieh]{rao2021dynamicvit}
Yongming Rao, Wenliang Zhao, Benlin Liu, Jiwen Lu, Jie Zhou, and Cho-Jui Hsieh.
\newblock Dynamicvit: Efficient vision transformers with dynamic token sparsification.
\newblock \emph{Advances in neural information processing systems}, 34:\penalty0 13937--13949, 2021.

\bibitem[Rasley et~al.(2020)Rasley, Rajbhandari, Ruwase, and He]{rasley2020deepspeed}
Jeff Rasley, Samyam Rajbhandari, Olatunji Ruwase, and Yuxiong He.
\newblock Deepspeed: System optimizations enable training deep learning models with over 100 billion parameters.
\newblock In \emph{Proceedings of the 26th ACM SIGKDD International Conference on Knowledge Discovery \& Data Mining}, pages 3505--3506, 2020.

\bibitem[Renggli et~al.(2022)Renggli, Pinto, Houlsby, Mustafa, Puigcerver, and Riquelme]{renggli2022learning}
Cedric Renggli, Andr{\'e}~Susano Pinto, Neil Houlsby, Basil Mustafa, Joan Puigcerver, and Carlos Riquelme.
\newblock Learning to merge tokens in vision transformers.
\newblock \emph{arXiv preprint arXiv:2202.12015}, 2022.

\bibitem[Rombach et~al.(2022)Rombach, Blattmann, Lorenz, Esser, and Ommer]{rombach2022high}
Robin Rombach, Andreas Blattmann, Dominik Lorenz, Patrick Esser, and Bj{\"o}rn Ommer.
\newblock High-resolution image synthesis with latent diffusion models.
\newblock In \emph{Proceedings of the IEEE/CVF Conference on Computer Vision and Pattern Recognition}, pages 10684--10695, 2022.

\bibitem[Saharia et~al.(2022)Saharia, Chan, Saxena, Li, Whang, Denton, Ghasemipour, Ayan, Mahdavi, Lopes, et~al.]{saharia2022photorealistic}
Chitwan Saharia, William Chan, Saurabh Saxena, Lala Li, Jay Whang, Emily Denton, Seyed Kamyar~Seyed Ghasemipour, Burcu~Karagol Ayan, S~Sara Mahdavi, Raphael~Gontijo Lopes, et~al.
\newblock Photorealistic text-to-image diffusion models with deep language understanding.
\newblock \emph{arXiv preprint arXiv:2205.11487}, 2022.

\bibitem[Salimans and Ho(2022)]{salimans2022progressive}
Tim Salimans and Jonathan Ho.
\newblock Progressive distillation for fast sampling of diffusion models.
\newblock \emph{arXiv preprint arXiv:2202.00512}, 2022.

\bibitem[Sauer et~al.(2025)Sauer, Lorenz, Blattmann, and Rombach]{sauer2025adversarial}
Axel Sauer, Dominik Lorenz, Andreas Blattmann, and Robin Rombach.
\newblock Adversarial diffusion distillation.
\newblock In \emph{European Conference on Computer Vision}, pages 87--103. Springer, 2025.

\bibitem[Shi et~al.(2023)Shi, Chen, Zhang, Liu, Xu, Wei, Chen, Zeng, and Su]{shi2023zero123++}
Ruoxi Shi, Hansheng Chen, Zhuoyang Zhang, Minghua Liu, Chao Xu, Xinyue Wei, Linghao Chen, Chong Zeng, and Hao Su.
\newblock Zero123++: a single image to consistent multi-view diffusion base model.
\newblock \emph{arXiv preprint arXiv:2310.15110}, 2023.

\bibitem[Singer et~al.(2022)Singer, Polyak, Hayes, Yin, An, Zhang, Hu, Yang, Ashual, Gafni, et~al.]{singer2022make}
Uriel Singer, Adam Polyak, Thomas Hayes, Xi Yin, Jie An, Songyang Zhang, Qiyuan Hu, Harry Yang, Oron Ashual, Oran Gafni, et~al.
\newblock Make-a-video: Text-to-video generation without text-video data.
\newblock \emph{arXiv preprint arXiv:2209.14792}, 2022.

\bibitem[Smith et~al.(2024)Smith, Saxena, and Saha]{smith2024todo}
Ethan Smith, Nayan Saxena, and Aninda Saha.
\newblock Todo: Token downsampling for efficient generation of high-resolution images.
\newblock \emph{arXiv preprint arXiv:2402.13573}, 2024.

\bibitem[So et~al.(2023)So, Lee, and Park]{so2023frdiff}
Junhyuk So, Jungwon Lee, and Eunhyeok Park.
\newblock Frdiff: Feature reuse for universal training-free acceleration of diffusion models.
\newblock \emph{arXiv preprint arXiv:2312.03517}, 2023.

\bibitem[So et~al.(2024)So, Lee, Ahn, Kim, and Park]{so2024temporal}
Junhyuk So, Jungwon Lee, Daehyun Ahn, Hyungjun Kim, and Eunhyeok Park.
\newblock Temporal dynamic quantization for diffusion models.
\newblock \emph{Advances in Neural Information Processing Systems}, 36, 2024.

\bibitem[Sohl-Dickstein et~al.(2015)Sohl-Dickstein, Weiss, Maheswaranathan, and Ganguli]{sohl2015deep}
Jascha Sohl-Dickstein, Eric Weiss, Niru Maheswaranathan, and Surya Ganguli.
\newblock Deep unsupervised learning using nonequilibrium thermodynamics.
\newblock In \emph{International conference on machine learning}, pages 2256--2265. PMLR, 2015.

\bibitem[Song et~al.(2024)Song, Wang, Chen, Wang, Guan, and Wang]{song2024less}
Dingjie Song, Wenjun Wang, Shunian Chen, Xidong Wang, Michael Guan, and Benyou Wang.
\newblock Less is more: A simple yet effective token reduction method for efficient multi-modal llms.
\newblock \emph{arXiv preprint arXiv:2409.10994}, 2024.

\bibitem[Song et~al.(2020)Song, Meng, and Ermon]{song2020denoising}
Jiaming Song, Chenlin Meng, and Stefano Ermon.
\newblock Denoising diffusion implicit models.
\newblock \emph{arXiv preprint arXiv:2010.02502}, 2020.

\bibitem[Song et~al.(2023)Song, Dhariwal, Chen, and Sutskever]{song2023consistency}
Yang Song, Prafulla Dhariwal, Mark Chen, and Ilya Sutskever.
\newblock Consistency models.
\newblock \emph{arXiv preprint arXiv:2303.01469}, 2023.

\bibitem[Tran et~al.(2024)Tran, Nguyen, Nguyen, Nguyen, Le, Xie, Sonntag, Zou, Nguyen, and Niepert]{tran2024accelerating}
Hoai-Chau Tran, Duy~MH Nguyen, Duy~M Nguyen, Trung-Tin Nguyen, Ngan Le, Pengtao Xie, Daniel Sonntag, James~Y Zou, Binh~T Nguyen, and Mathias Niepert.
\newblock Accelerating transformers with spectrum-preserving token merging.
\newblock \emph{arXiv preprint arXiv:2405.16148}, 2024.

\bibitem[Wang et~al.(2024{\natexlab{a}})Wang, Dedhia, and Jha]{wang2024zero}
Hongjie Wang, Bhishma Dedhia, and Niraj~K Jha.
\newblock Zero-tprune: Zero-shot token pruning through leveraging of the attention graph in pre-trained transformers.
\newblock In \emph{Proceedings of the IEEE/CVF Conference on Computer Vision and Pattern Recognition}, pages 16070--16079, 2024{\natexlab{a}}.

\bibitem[Wang et~al.(2024{\natexlab{b}})Wang, Liu, Kang, Li, Lin, Jha, and Liu]{wang2024attention}
Hongjie Wang, Difan Liu, Yan Kang, Yijun Li, Zhe Lin, Niraj~K Jha, and Yuchen Liu.
\newblock Attention-driven training-free efficiency enhancement of diffusion models.
\newblock In \emph{Proceedings of the IEEE/CVF Conference on Computer Vision and Pattern Recognition}, pages 16080--16089, 2024{\natexlab{b}}.

\bibitem[Wang et~al.(2024{\natexlab{c}})Wang, Shang, Yuan, Wu, Yan, and Yan]{wang2024quest}
Haoxuan Wang, Yuzhang Shang, Zhihang Yuan, Junyi Wu, Junchi Yan, and Yan Yan.
\newblock Quest: Low-bit diffusion model quantization via efficient selective finetuning.
\newblock \emph{arXiv preprint arXiv:2402.03666}, 2024{\natexlab{c}}.

\bibitem[Wang et~al.(2024{\natexlab{d}})Wang, Chen, Li, Mi, and Zhu]{wang2024sparsedm}
Kafeng Wang, Jianfei Chen, He Li, Zhenpeng Mi, and Jun Zhu.
\newblock Sparsedm: Toward sparse efficient diffusion models.
\newblock \emph{arXiv preprint arXiv:2404.10445}, 2024{\natexlab{d}}.

\bibitem[Wang and Yang(2024)]{wang2024efficient}
Yancheng Wang and Yingzhen Yang.
\newblock Efficient visual transformer by learnable token merging.
\newblock \emph{arXiv preprint arXiv:2407.15219}, 2024.

\bibitem[Wang et~al.(2024{\natexlab{e}})Wang, Zhang, Zheng, and Jin]{wang2024high}
Yibin Wang, Weizhong Zhang, Jianwei Zheng, and Cheng Jin.
\newblock High-fidelity person-centric subject-to-image synthesis.
\newblock In \emph{Proceedings of the IEEE/CVF Conference on Computer Vision and Pattern Recognition}, pages 7675--7684, 2024{\natexlab{e}}.

\bibitem[Wang et~al.(2004)Wang, Bovik, Sheikh, and Simoncelli]{wang2004image}
Zhou Wang, Alan~C Bovik, Hamid~R Sheikh, and Eero~P Simoncelli.
\newblock Image quality assessment: from error visibility to structural similarity.
\newblock \emph{IEEE transactions on image processing}, 13\penalty0 (4):\penalty0 600--612, 2004.

\bibitem[Wei et~al.(2023)Wei, Ye, Zhang, Tang, and Liang]{wei2023joint}
Siyuan Wei, Tianzhu Ye, Shen Zhang, Yao Tang, and Jiajun Liang.
\newblock Joint token pruning and squeezing towards more aggressive compression of vision transformers.
\newblock In \emph{Proceedings of the IEEE/CVF Conference on Computer Vision and Pattern Recognition}, pages 2092--2101, 2023.

\bibitem[Wimbauer et~al.(2024)Wimbauer, Wu, Schoenfeld, Dai, Hou, He, Sanakoyeu, Zhang, Tsai, Kohler, et~al.]{wimbauer2024cache}
Felix Wimbauer, Bichen Wu, Edgar Schoenfeld, Xiaoliang Dai, Ji Hou, Zijian He, Artsiom Sanakoyeu, Peizhao Zhang, Sam Tsai, Jonas Kohler, et~al.
\newblock Cache me if you can: Accelerating diffusion models through block caching.
\newblock In \emph{Proceedings of the IEEE/CVF Conference on Computer Vision and Pattern Recognition}, pages 6211--6220, 2024.

\bibitem[Wu et~al.(2023)Wu, Zeng, Wang, and Chen]{wu2023ppt}
Xinjian Wu, Fanhu Zeng, Xiudong Wang, and Xinghao Chen.
\newblock Ppt: Token pruning and pooling for efficient vision transformers.
\newblock \emph{arXiv preprint arXiv:2310.01812}, 2023.

\bibitem[Xu et~al.(2022{\natexlab{a}})Xu, De~Mello, Liu, Byeon, Breuel, Kautz, and Wang]{xu2022groupvit}
Jiarui Xu, Shalini De~Mello, Sifei Liu, Wonmin Byeon, Thomas Breuel, Jan Kautz, and Xiaolong Wang.
\newblock Groupvit: Semantic segmentation emerges from text supervision.
\newblock In \emph{Proceedings of the IEEE/CVF Conference on Computer Vision and Pattern Recognition}, pages 18134--18144, 2022{\natexlab{a}}.

\bibitem[Xu et~al.(2022{\natexlab{b}})Xu, Zhang, Zhang, Sheng, Li, Dong, Zhang, Xu, and Sun]{xu2022evo}
Yifan Xu, Zhijie Zhang, Mengdan Zhang, Kekai Sheng, Ke Li, Weiming Dong, Liqing Zhang, Changsheng Xu, and Xing Sun.
\newblock Evo-vit: Slow-fast token evolution for dynamic vision transformer.
\newblock In \emph{Proceedings of the AAAI Conference on Artificial Intelligence}, pages 2964--2972, 2022{\natexlab{b}}.

\bibitem[Xue et~al.(2024)Xue, Song, Guo, Liu, Zong, Liu, and Luo]{xue2024raphael}
Zeyue Xue, Guanglu Song, Qiushan Guo, Boxiao Liu, Zhuofan Zong, Yu Liu, and Ping Luo.
\newblock Raphael: Text-to-image generation via large mixture of diffusion paths.
\newblock \emph{Advances in Neural Information Processing Systems}, 36, 2024.

\bibitem[Yang et~al.(2023)Yang, Zhou, Feng, and Wang]{yang2023diffusion}
Xingyi Yang, Daquan Zhou, Jiashi Feng, and Xinchao Wang.
\newblock Diffusion probabilistic model made slim.
\newblock In \emph{Proceedings of the IEEE/CVF Conference on computer vision and pattern recognition}, pages 22552--22562, 2023.

\bibitem[Yin et~al.(2022)Yin, Vahdat, Alvarez, Mallya, Kautz, and Molchanov]{yin2022vit}
Hongxu Yin, Arash Vahdat, Jose~M Alvarez, Arun Mallya, Jan Kautz, and Pavlo Molchanov.
\newblock A-vit: Adaptive tokens for efficient vision transformer.
\newblock In \emph{Proceedings of the IEEE/CVF conference on computer vision and pattern recognition}, pages 10809--10818, 2022.

\bibitem[Zeng et~al.(2022)Zeng, Jin, Liu, Qian, Luo, Ouyang, and Wang]{zeng2022not}
Wang Zeng, Sheng Jin, Wentao Liu, Chen Qian, Ping Luo, Wanli Ouyang, and Xiaogang Wang.
\newblock Not all tokens are equal: Human-centric visual analysis via token clustering transformer.
\newblock In \emph{Proceedings of the IEEE/CVF Conference on Computer Vision and Pattern Recognition}, pages 11101--11111, 2022.

\bibitem[Zhang et~al.(2018)Zhang, Isola, Efros, Shechtman, and Wang]{zhang2018unreasonable}
Richard Zhang, Phillip Isola, Alexei~A Efros, Eli Shechtman, and Oliver Wang.
\newblock The unreasonable effectiveness of deep features as a perceptual metric.
\newblock In \emph{Proceedings of the IEEE conference on computer vision and pattern recognition}, pages 586--595, 2018.

\bibitem[Zhao et~al.(2023{\natexlab{a}})Zhao, Zheng, Wang, Lan, and Yang]{zhao2023magicfusion}
Jing Zhao, Heliang Zheng, Chaoyue Wang, Long Lan, and Wenjing Yang.
\newblock Magicfusion: Boosting text-to-image generation performance by fusing diffusion models.
\newblock In \emph{Proceedings of the IEEE/CVF International Conference on Computer Vision}, pages 22592--22602, 2023{\natexlab{a}}.

\bibitem[Zhao et~al.(2024{\natexlab{a}})Zhao, Han, Tang, Wang, Song, Huang, Wang, and You]{zhao2024dynamic}
Wangbo Zhao, Yizeng Han, Jiasheng Tang, Kai Wang, Yibing Song, Gao Huang, Fan Wang, and Yang You.
\newblock Dynamic diffusion transformer.
\newblock \emph{arXiv preprint arXiv:2410.03456}, 2024{\natexlab{a}}.

\bibitem[Zhao et~al.(2024{\natexlab{b}})Zhao, Jin, Wang, and You]{zhao2024real}
Xuanlei Zhao, Xiaolong Jin, Kai Wang, and Yang You.
\newblock Real-time video generation with pyramid attention broadcast.
\newblock \emph{arXiv preprint arXiv:2408.12588}, 2024{\natexlab{b}}.

\bibitem[Zhao et~al.(2023{\natexlab{b}})Zhao, Xu, Xiao, and Hou]{zhao2023mobilediffusion}
Yang Zhao, Yanwu Xu, Zhisheng Xiao, and Tingbo Hou.
\newblock Mobilediffusion: Subsecond text-to-image generation on mobile devices.
\newblock \emph{arXiv preprint arXiv:2311.16567}, 2023{\natexlab{b}}.

\bibitem[Zheng et~al.(2024)Zheng, Peng, Yang, Shen, Li, Liu, Zhou, Li, and You]{opensora}
Zangwei Zheng, Xiangyu Peng, Tianji Yang, Chenhui Shen, Shenggui Li, Hongxin Liu, Yukun Zhou, Tianyi Li, and Yang You.
\newblock Open-sora: Democratizing efficient video production for all, 2024.

\bibitem[Zong et~al.(2022)Zong, Li, Song, Wang, Qiao, Leng, and Liu]{zong2022self}
Zhuofan Zong, Kunchang Li, Guanglu Song, Yali Wang, Yu Qiao, Biao Leng, and Yu Liu.
\newblock Self-slimmed vision transformer.
\newblock In \emph{European Conference on Computer Vision}, pages 432--448. Springer, 2022.

\end{thebibliography}
}
\appendix
\clearpage
\setcounter{page}{1}
\maketitlesupplementary

In~\cref{sec:additional_results}, we present more qualitative comparisons, empirical evidence on the relationship between CFG and token importance, results on consistency models, results on combining our method with orthogonal diffusion acceleration techniques, and experiments with varying diffusion inference steps. In~\cref{sec:additional_exp_settings}, we provide more details about our experimental settings. In~\cref{sec:limits}, we discuss the limitations of our method. In~\cref{sec:prompts}, we provide the prompts used to generate qualitative results. We also include a supplementary video for comparisons on text-to-video generation.

\section{Additional Results}
\label{sec:additional_results}

\begin{figure*}[t]
\begin{center}
\includegraphics[width=1\textwidth]{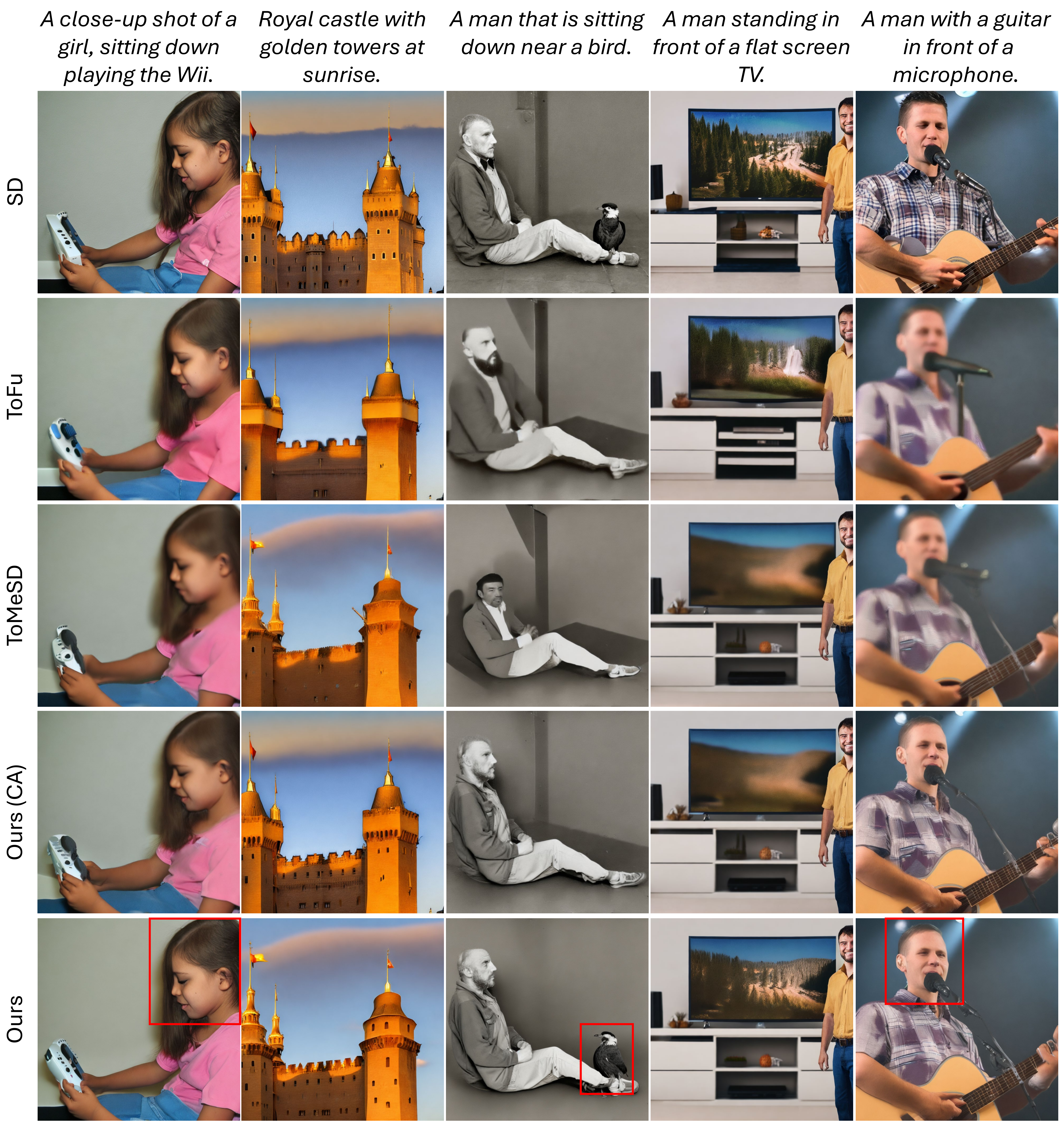}
\end{center}
\caption{\textbf{Additional comparison of text-to-image generation.} The first row shows results from Stable Diffusion (SD)~\cite{rombach2022high}, while the subsequent rows show SD combined with ToFu~\cite{kim2024token}, ToMeSD~\cite{Bolya2023TokenMF}, our method using cross-attention (CA) map, and our method using classifier-free guidance. The token merging ratio is 0.7. Our method outputs finer details, as highlighted in
red boxes. Notably, the variant of our method that utilizes the cross-attention map also achieves better generation details compared to baseline methods, demonstrating the generalization ability of our method. Best viewed with zoom-in for clarity.}
\label{fig:t2i_supp}
\end{figure*}

\begin{figure*}[t]
\begin{center}
\includegraphics[width=0.9\textwidth]{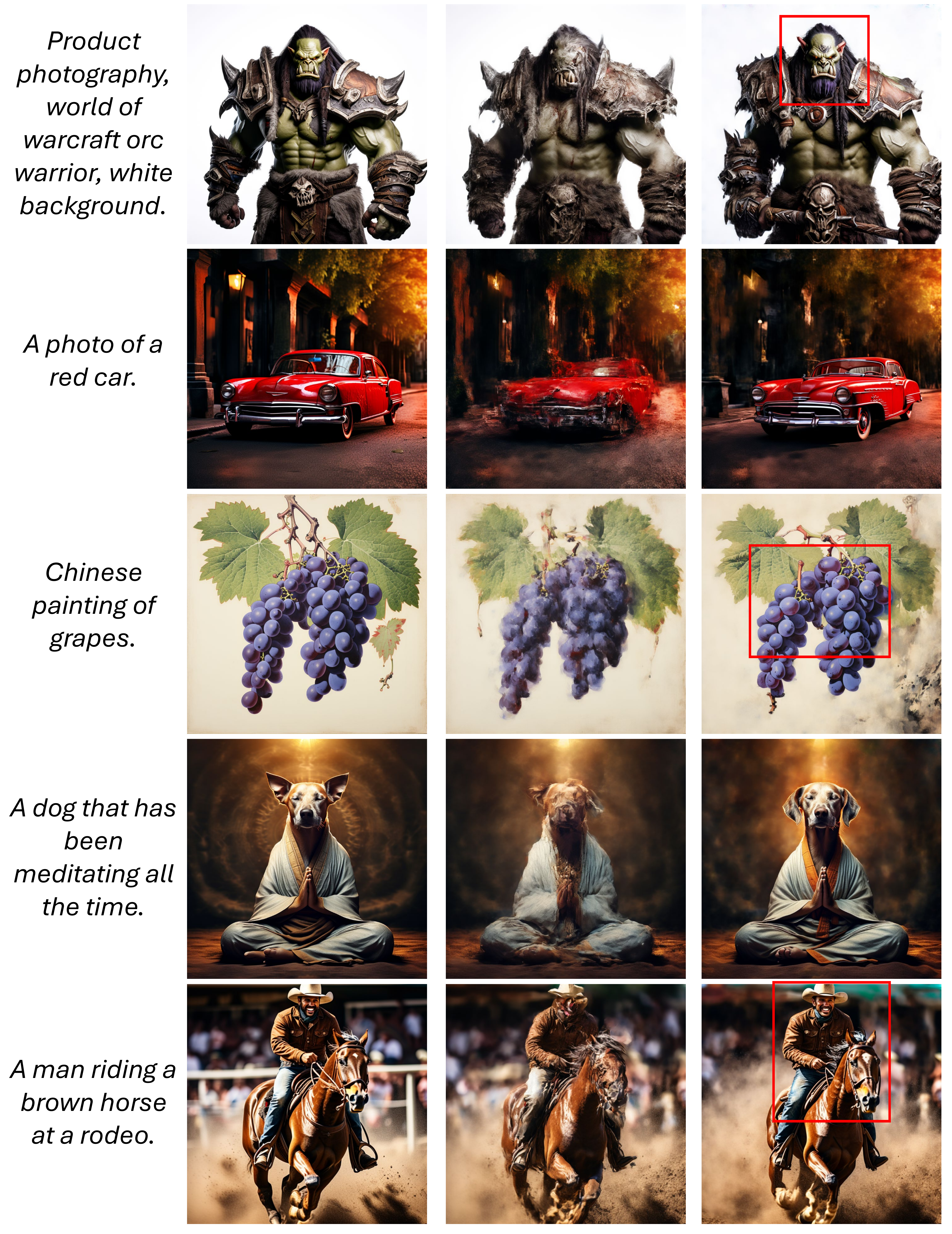}
\makebox[0.16\linewidth]{}%
\makebox[0.24\linewidth]{\centering (a) PixArt-$\alpha$}%
\makebox[0.24\linewidth]{\centering (b) ToMeSD}%
\makebox[0.24\linewidth]{\centering (c) Ours}
\end{center}
\vspace{-5mm}
\caption{\textbf{Additional comparison for token merging applied to diffusion transformer.} We apply ToMeSD~\cite{Bolya2023TokenMF} and our token merging method to PixArt-$\alpha$~\cite{chen2023pixartalpha} for text-to-image synthesis, using a token merging ratio of 0.4. We highlight our generation details with red boxes. Best viewed with zoom-in for clarity.}
\label{fig:pixart_supp}
\end{figure*}

\paragraph{Additional Qualitative Results.} 
In~\cref{fig:t2i_supp}, we provide additional qualitative comparisons between ToFu~\cite{kim2024token}, ToMeSD~\cite{Bolya2023TokenMF}, our token merging method, and the variant of our method using cross-attention maps as importance signals. 
Additional visual comparisons for token merging applied to diffusion transformer are shown in ~\cref{fig:pixart_supp}. 
Additional results for multi-view diffusion are presented in ~\cref{fig:mv_supp}.
In the supplementary video, we include comparisons on text-to-video generation, using AnimateDiff~\cite{guo2023animatediff} as the base diffusion model and a merging ratio of 0.7. Furthermore, we provide visual comparisons between ToMeSD~\cite{Bolya2023TokenMF} and our method across various merging ratios for text-to-image generation in~\cref{fig:ratios}.

\paragraph{Token Importance via Classifier-Free Guidance.}
The absolute value of classifier-free guidance (CFG) can be interpreted as a token-level saliency measure, highlighting the tokens that play a crucial role in steering the output toward the given prompt or condition. Empirically, as shown in~\cref{tab:cfg_rebuttal}, removing the top 30\% of high-CFG tokens leads to a significant degradation in generation results, whereas removing the bottom 30\% has little impact.

\begin{table}[htb]
    \centering
    \begin{tabular}{cccc}
    \toprule
     & No pruning & Top 30\% & Bottom 30\% \\
     \midrule
    FID $\downarrow$ & 11.88 & 15.48 & 12.78 \\
    CLIP $\uparrow$ & 31.83 & 31.58 & 31.88 \\
    \bottomrule
    \end{tabular}
    \vspace{-1mm}
    \caption{Comparison of pruning (dropping) the top 30\% of tokens with the highest CFG values versus the bottom 30\% during text-to-image generation using Stable Diffusion~\cite{rombach2022high}.}
    \label{tab:cfg_rebuttal}
\end{table}

\paragraph{Results on the Consistency Model.} 
Consistency models~\cite{song2023consistency, luo2023latent, luo2023lcm} typically distill classifier-free guidance (CFG), making the explicit guidance term inaccessible, as they approximate the final guided noise function within a single forward pass. However, our method is not limited to CFG and can leverage any reliable per-token importance signal. As shown in~\cref{tab:consistency_model}, our token merging, using cross-attention maps as importance signals, remains effective and outperforms the baseline model when applied to the latent consistency model~\cite{luo2023lcm}.

\begin{table}[htb]
    \centering
    \begin{tabular}{ccccccc}
    \toprule
     \multirow{2}{*}{$r$} & \multicolumn{2}{c}{FID $\downarrow$} & \multicolumn{2}{c}{CLIP $\uparrow$} & Time & Mem. \\
    \cmidrule(lr){2-3} \cmidrule(lr){4-5}
    & ToMe. & Ours & ToMe. & Ours & (s) $\downarrow$ & (GB)$\downarrow$ \\
    \midrule
    0    & \multicolumn{2}{c}{25.38} & \multicolumn{2}{c}{31.05} & 0.65 & 6.75 \\
    0.30 & 25.63 & 25.63 & 31.03 & 31.03 & 0.60 & 4.22 \\
    0.50 & 27.80 & \textbf{27.51} & 30.98 & \textbf{30.99} & 0.51 & 3.56\\
    0.60 & 30.05 & \textbf{29.61} & 30.90 & \textbf{30.92} & 0.50 & 3.37\\
    0.70 & 35.01 & \textbf{32.49} & 30.59 & \textbf{30.77} & 0.48 & 3.21 \\
    0.75 & 43.28 & \textbf{36.66} & 30.38 & \textbf{30.60} & 0.46 & 3.12 \\
    \bottomrule
    \end{tabular}
    \vspace{-1mm}
    \caption{\textbf{Results on the Consistency Model.} We show the comparison of token merging methods applied to a 4-step latent consistency model (LCM\_Dreamshaper\_v7)~\cite{luo2023lcm} across various merging ratios $r$.}
    \label{tab:consistency_model}
\end{table}

\paragraph{Combination with Orthogonal Diffusion Acceleration Methods.} In~\cref{tab:orthogonal_methods}, we demonstrate that our method can be combined with orthogonal diffusion acceleration methods~\cite{ma2024deepcache, li2023faster, so2023frdiff} to further accelerate inference while preserving generation quality.

\begin{table}[htb]
    \centering
    \begin{tabular}{lcccc}
    \toprule
     & \multirow{2}{*}{FID $\downarrow$} & \multirow{2}{*}{CLIP $\uparrow$} & Time & Mem. \\
     & & & (s) $\downarrow$ & (GB) $\downarrow$ \\
    \midrule
    Ours     & 16.22 & 31.79 & 5.8 & 3.55 \\
    + DeepCache~\cite{ma2024deepcache}   & 15.46 & 31.81 & 2.6 & 3.56 \\
    + FasterDiff.~\cite{li2023faster} & 12.48 & 31.80 & 4.2 & 6.33 \\ 
    + FRDiff~\cite{so2023frdiff}  & 15.25 & 31.83 & 3.5 & 3.59 \\
    \bottomrule
    \end{tabular}
    \vspace{-1mm}
    \caption{\textbf{Combination with Diffusion Acceleration Methods}. We show text-to-image generation results by integrating our method with orthogonal diffusion acceleration techniques, using Stable Diffusion~\cite{rombach2022high} as the base model and a merging ratio of 0.7.}
    \label{tab:orthogonal_methods}
\end{table}

\paragraph{Number of Diffusion Inference Steps} 
In~\cref{tab:steps}, we compare ToMeSD~\cite{Bolya2023TokenMF} and our method across different numbers of diffusion inference steps for text-to-image generation, demonstrating that our method consistently outperforms ToMeSD.

\begin{table}[htb]
\centering
\begin{tabular}{ccccc}
\toprule
 \multirow{2}{*}{$T$} & \multicolumn{2}{c}{FID $\downarrow$} & \multicolumn{2}{c}{CLIP $\uparrow$} \\
\cmidrule(lr){2-3} \cmidrule(lr){4-5}
& ToMeSD & Ours & ToMeSD & Ours \\
\midrule
20 & 18.57 & \textbf{17.03} & 31.77 & \textbf{31.80} \\
30 & 17.82 & \textbf{16.51} & 31.78 & \textbf{31.80} \\
50 & 17.46 & \textbf{16.22} & 31.78 & \textbf{31.79} \\
\bottomrule
\end{tabular}
\vspace{-1mm}
\caption{\textbf{Number of Diffusion Inference Steps.} We compare ToMeSD~\cite{Bolya2023TokenMF} and our token merging method for the text-to-image generation task with different diffusion inference steps, using Stable Diffusion~\cite{rombach2022high} as the base model and a merging ratio of 0.7.}
\label{tab:steps}
\end{table}

\section{Additional Experimental Settings}
\label{sec:additional_exp_settings}

\paragraph{Metrics.} 
We use the \textit{deepspeed}~\cite{rasley2020deepspeed} library to estimate TFLOPs, the \textit{clean-fid}~\cite{parmar2021cleanfid} library to calculate FID scores, and the \textit{openai/clip-vit-base-patch16} model from OpenAI-CLIP~\cite{radford2021learning} to calculate CLIP scores.

\paragraph{Additional Implementation Details.}
For text-to-image generation using Stable Diffusion~\cite{rombach2022high}, the diffusion process consists of 50 sampling steps, with the CFG scale set to 7.5. 
For PixArt-$\alpha$~\cite{chen2023pixartalpha}, we perform diffusion sampling for 20 steps, with a CFG scale of 4.5.
When computing token similarity for token merging in PixArt-$\alpha$, we find that using pixel location distance of tokens yields better results than feature similarity, and we adopt this approach.
For multi-view diffusion using Zero123++~\cite{shi2023zero123++}, the process involves 50 sampling steps, with the CFG scale set to 4. 
For video diffusion using AnimateDiff~\cite{guo2023animatediff}, the sampling consists of 30 steps, with a CFG scale of 7.5. 
To ensure fairness, these settings are consistently applied to both our method and the baselines. 
For the ablation study investigating the use of cross-attention maps as importance signals, we utilize the averaged attention map from the final model block in Stable Diffusion.

\paragraph{Details on Multi-view Diffusion.} 
The base multi-view diffusion model used in our experiments is Zero123++~\cite{shi2023zero123++}, which fine-tunes Stable Diffusion 2~\cite{rombach2022high} to generate six novel views from an input image. During denoising, the model appends the self-attention key and value matrices from the reference input image to the attention layers for conditioning. The novel view poses are defined by a fixed
set of absolute elevation and relative azimuth angles. Specifically, the elevation and azimuth angles (in degrees) are set as follows: (30, 30), (-20, 90), (30, 150), (-20, 210), (30, 270), (-20, 330).  We consistently use this sequence of novel views to present visual results in multi-view diffusion experiments.

\paragraph{Preserving Structure in Early Time-steps.}
Prior work~\cite{kim2024token} suggests that token pruning (directly dropping tokens) could help preserve structural details. Building on this observation, we found that incorporate token pruning during the early diffusion steps, followed by token merging in later steps, improves generation results. Specifically, we apply token pruning during the first 6, 10, and 4 diffusion inference steps for image, multi-view, and video generation, respectively. 
Furthermore, the low token variance in the early steps~\cite{wang2024attention} reduces the effectiveness of classifier-free guidance in identifying important tokens. To mitigate this, during these initial steps that involve token pruning, we randomly select one token from each $2 \times 2$ region of the feature map as the destination token.

In~\cref{tab:prune} and~\cref{fig:prune}, we compare the results of ToMeSD~\cite{Bolya2023TokenMF} under two scenarios: (1) applying token pruning during the early diffusion steps followed by token merging, and (2) using token merging throughout all diffusion steps. The results show that token pruning in the early steps more effectively preserves the generation layout. In \cref{tab:prune_steps}, we show that pruning tokens during the first 5-20\% denoising steps consistently improves performance across different tasks, highlighting its robustness.

\begin{table}[htb]
\centering
\begin{tabular}{ccccc}
\toprule
 \multirow{2}{*}{$r$} & \multicolumn{2}{c}{FID $\downarrow$} & \multicolumn{2}{c}{CLIP $\uparrow$} \\
\cmidrule(lr){2-3} \cmidrule(lr){4-5}
     & w/o pr. & w/ pr. & w/o pr. & w/ pr. \\
\midrule
0.10 & 11.75 & \textbf{11.72} & 31.81 & \textbf{31.82} \\
0.30 & \textbf{12.16} & 12.20 & 31.82 & 31.82 \\
0.50 & \textbf{13.49} & 13.50 & 31.79 & 31.79 \\
0.60 & 14.81 & 14.81 & 31.79 & \textbf{31.80} \\
0.70 & 17.51 & \textbf{17.46} & 31.76 & \textbf{31.78} \\
0.75 & 21.05 & \textbf{20.89} & 31.69 & \textbf{31.71} \\
\bottomrule
\end{tabular}
\vspace{-1mm}
\caption{Ablation studies on token pruning in early diffusion inference steps. We compare the results of ToMeSD~\cite{Bolya2023TokenMF} with token pruning in early diffusion inference steps followed by token merging, versus using token merging for all steps. We evaluate using the text-to-image generation task with Stable Diffusion~\cite{rombach2022high} as the base model across various token merging ratios $r$.}
\label{tab:prune}
\end{table}

\begin{table}[htb]
    \centering
    \begin{tabular}{lc|ccc}
    \toprule
    \multirow{2}{*}{Metric} & \multicolumn{4}{c}{Pruning Steps (m\%)} \\
    \cmidrule(lr){2-5}
     & 0\% & 5\% & 10\% & 20\% \\
    \midrule
    \multicolumn{5}{l}{\textit{Image}} \\
    \quad FID $\downarrow$ & 16.27 & 16.28 & 16.22 & \textbf{16.13} \\
    \quad CLIP $\uparrow$ & 31.77 & 31.79 & \textbf{31.79} & 31.79 \\
    \addlinespace
    \multicolumn{5}{l}{\textit{Multi-View}} \\
    \quad PSNR $\uparrow$ & 14.73 & \textbf{14.95} & 14.75 & 14.80 \\
    \quad SSIM $\uparrow$ & 0.783 & 0.782 & \textbf{0.787} & 0.785 \\
    \quad LPIPS $\downarrow$ & 0.279 & 0.269 & \textbf{0.268} & 0.274 \\
    \addlinespace
    \multicolumn{5}{l}{\textit{Video}} \\
    \quad Score $\uparrow$ & 79.36 & 79.59 & 79.63 & \textbf{79.66} \\
    \bottomrule
    \end{tabular}
    \caption{We apply token pruning to the first \( m\%\) denoising steps (merging ratio = 0.7) and evaluate on three generation tasks.}
    \label{tab:prune_steps}
\end{table}

\section{Discussion and Limitations}
\label{sec:limits}
Our method demonstrates broad applicability across diffusion models. For multi-guidance scenarios (\eg, InstructPix2Pix~\cite{brooks2023instructpix2pix}), weighted averaging of importance signals based on user preferences could be beneficial. For step-distilled diffusion models~\cite{meng2023distillation}, which already distills classifier-free guidance into the final model, our approach can still be applied by utilizing alternative importance signals, such as attention maps. However, an interesting direction for future research could involve refining the distillation process to enable the model to predict an additional output: a classifier-free guidance map, which could then be used for better token merging or other innovative applications.

\section{Prompts}
\label{sec:prompts}

We provide the prompts used to generate the qualitative results shown in the paper but not included in the figures.

Text prompts corresponding to the text-to-image generations 
in Figure 5 of the main paper:
\begin{itemize}[left=\parindent]
\item \textit{Elegant teacup with a delicate floral pattern}

\item \textit{Young musician playing guitar on stage}

\item \textit{Colorful butterfly with wings fully spread}

\end{itemize}

Text prompts corresponding to the text-to-image generations 
in Figure 6 of the main paper:
\begin{itemize}[left=\parindent]
\item \textit{A cute cat}

\item \textit{A real beautiful face}

\item \textit{A small cactus with a happy face in the Sahara desert}

\end{itemize}

Text prompts corresponding to the text-to-video generations 
in Figure 8 of the main paper:
\begin{itemize}[left=\parindent]
\item \textit{Tower}

\item \textit{A jellyfish floating through the ocean, with bioluminescent tentacles}

\item \textit{In a still frame, the ornate Victorian streetlamp stands solemnly, adorned with intricate ironwork and stained glass panels}

\end{itemize}

In~\cref{fig:image_prompts1}, we show image prompts corresponding to the image-conditioned multi-view generations 
in Figure 7 of the main paper.

\begin{figure*}[t]
\begin{center}
\includegraphics[width=0.85\textwidth]{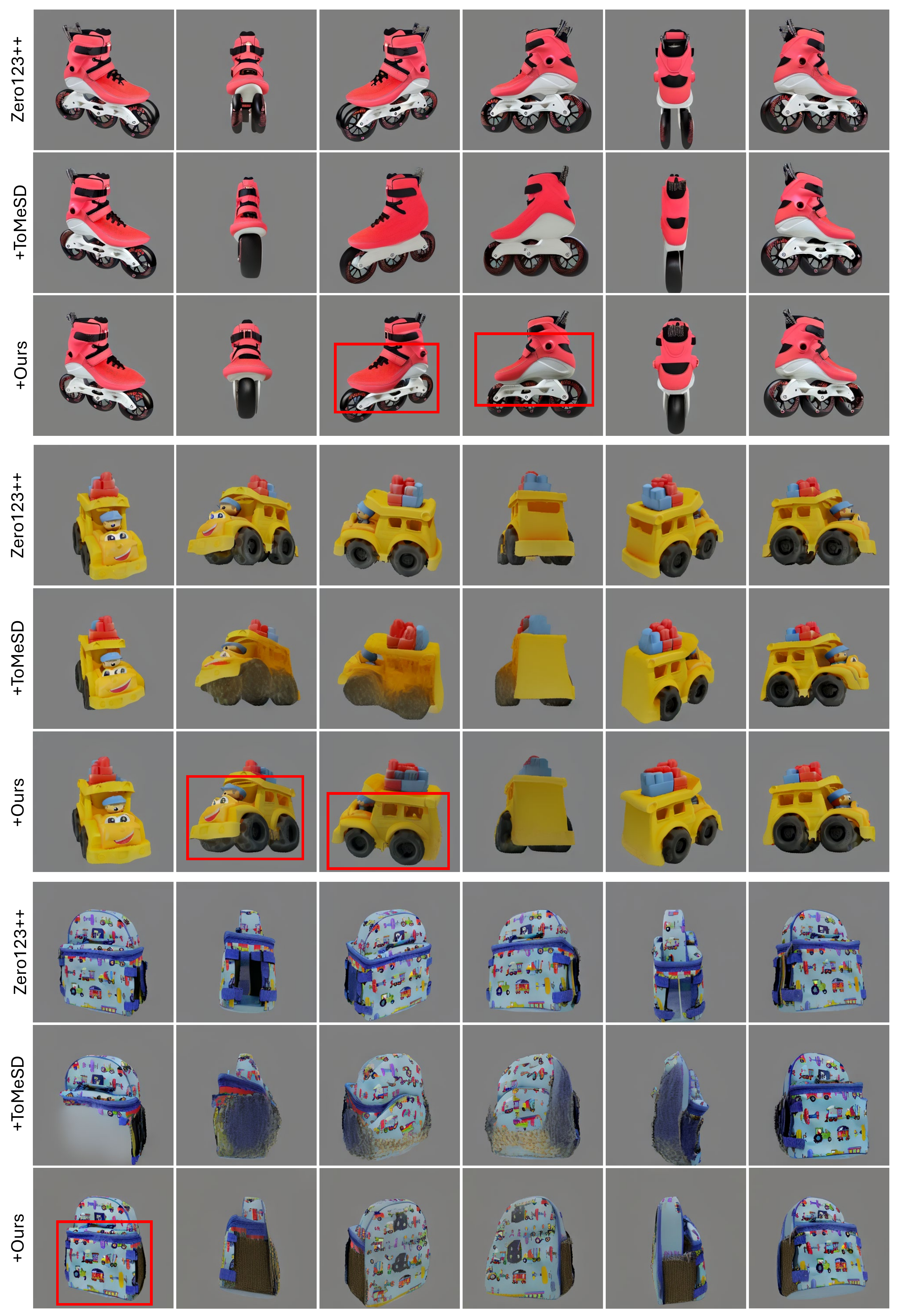}
\end{center}
\vspace{-7mm}
\caption{\textbf{Additional qualitative comparison of multi-view diffusion.} Token merging is applied to the multi-view diffusion model, Zero123++~\cite{shi2023zero123++}, with merging ratio as 0.6. Our method outputs finer details, as highlighted in
red boxes. Best viewed with zoom-in.}
\label{fig:mv_supp}
\end{figure*}

\begin{figure*}[t]
\begin{center}
\includegraphics[width=1\textwidth]{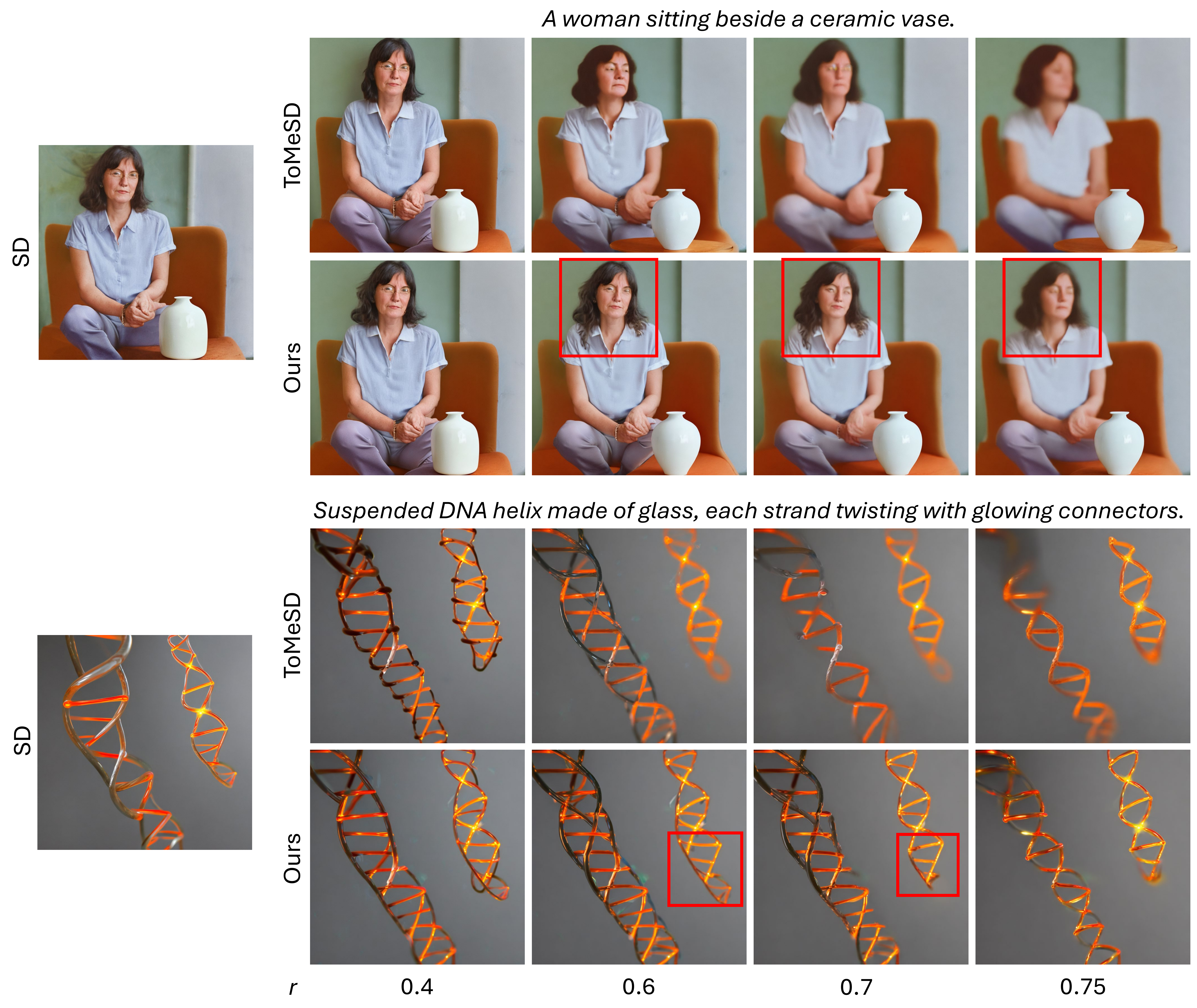}
\end{center}
\caption{We provide an additional comparison between ToMeSD~\cite{Bolya2023TokenMF} and our method when applied to Stable Diffusion~\cite{rombach2022high} across various merging ratios $r$. For reference, the results of Stable Diffusion without token merging are shown on the left. Our method outputs finer details, as highlighted in
red boxes. Best viewed with zoom-in for clarity.}
\label{fig:ratios}
\end{figure*}

\begin{figure*}[t]
\begin{center}
\includegraphics[width=0.85\textwidth]{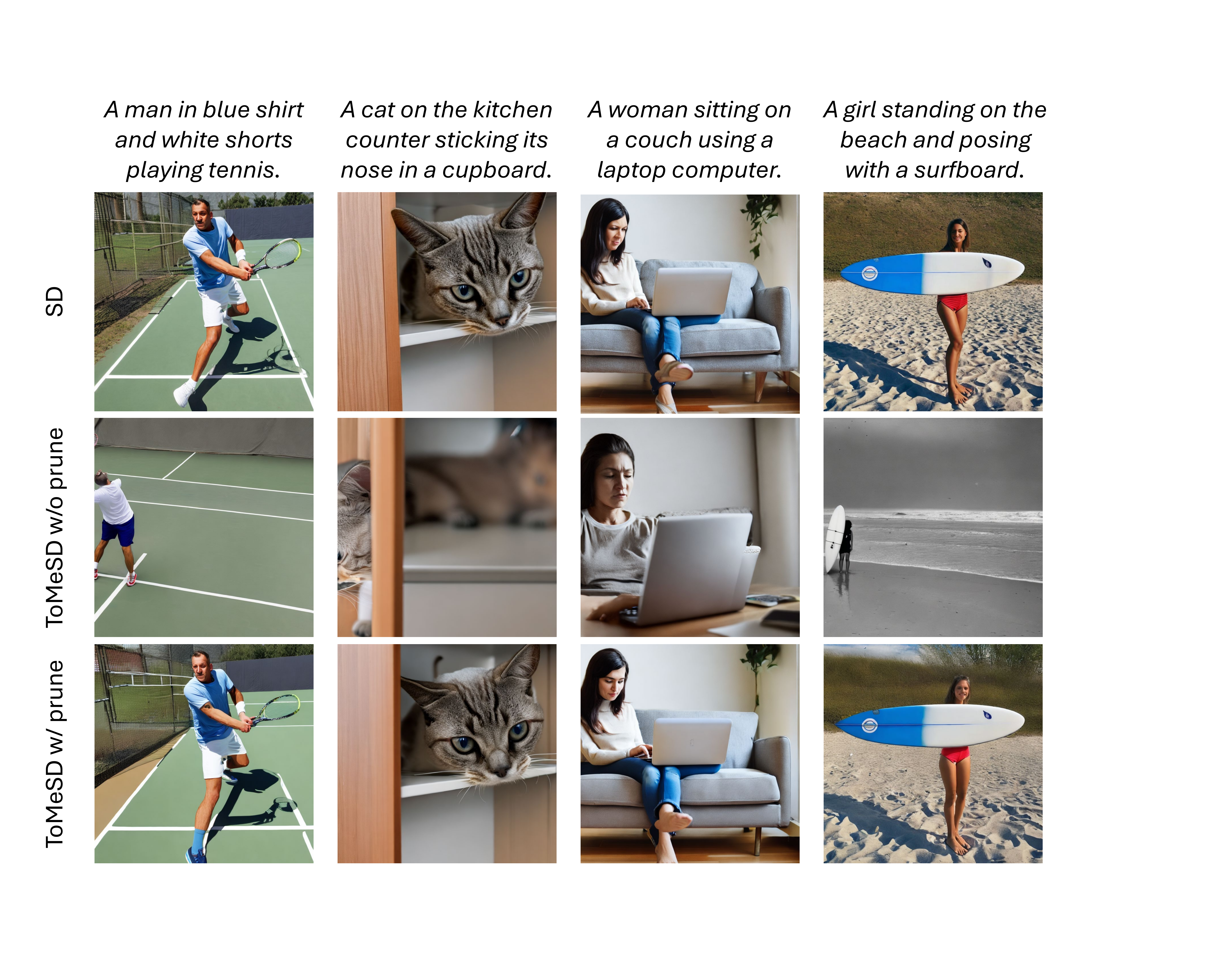}
\end{center}
\caption{We compare the results of ToMeSD~\cite{Bolya2023TokenMF} with token pruning in early diffusion inference steps followed by token merging (w/ prune), versus using token merging for all steps (w/o prune). We use Stable Diffusion~\cite{rombach2022high} as the base model and a merging ratio of 0.6.}
\label{fig:prune}
\end{figure*}

\begin{figure}[t]
\begin{center}
\includegraphics[width=0.8\linewidth]{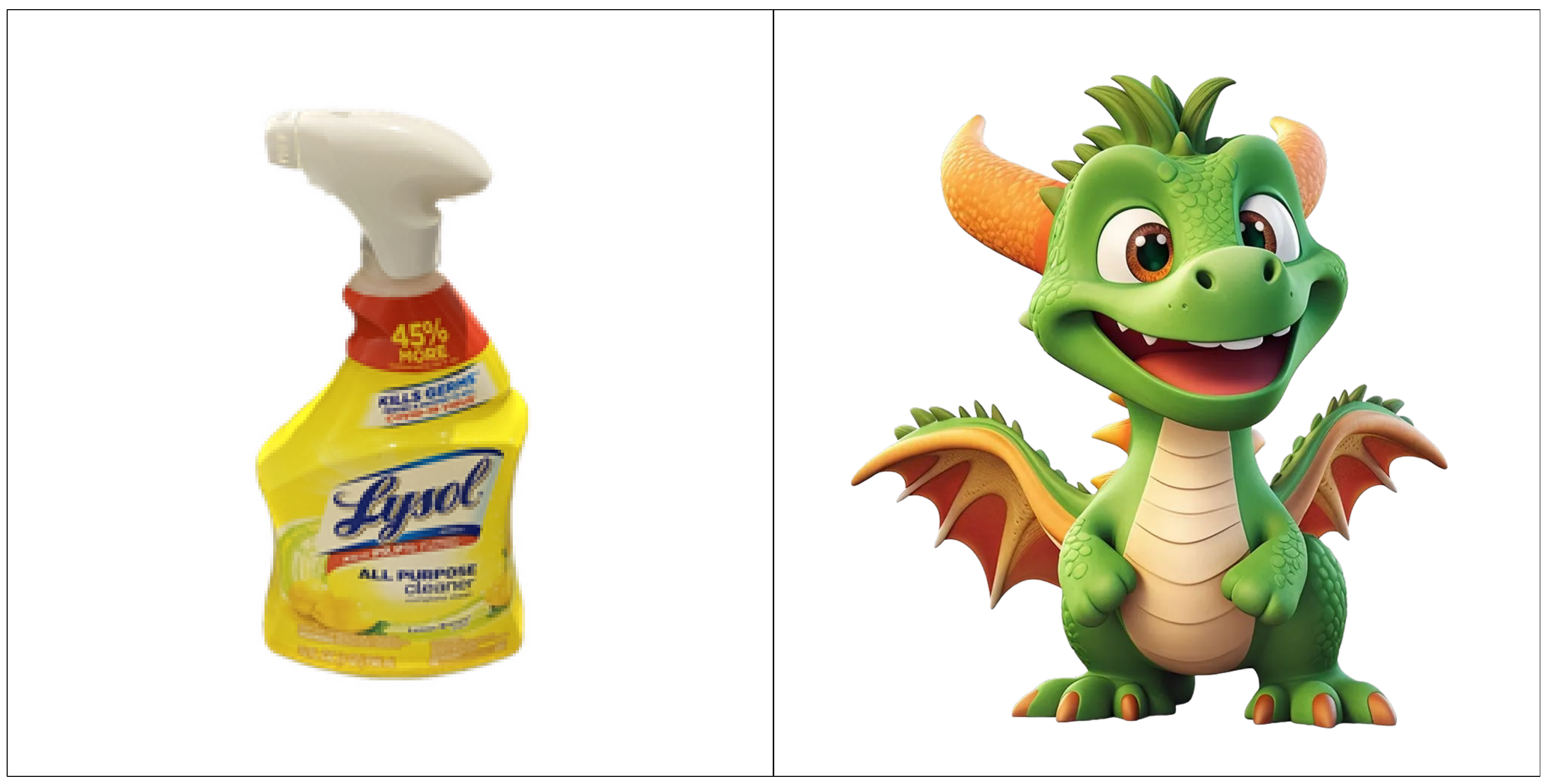}
\end{center}
\caption{Image prompts 
for Figure 7 of the main paper.
}
\label{fig:image_prompts1}
\end{figure}

\begin{figure}[t]
\begin{center}
\includegraphics[width=1\linewidth]{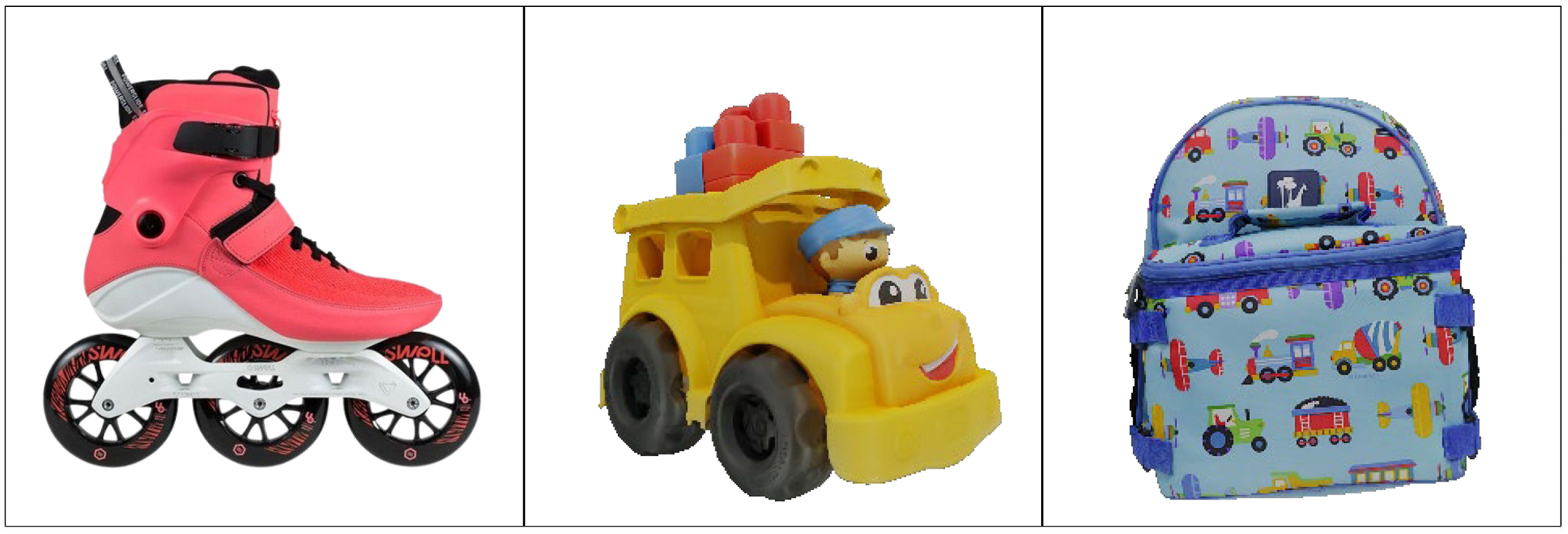}
\end{center}
\caption{Image prompts for~\cref{fig:mv_supp}.}
\label{fig:image_prompts2}
\end{figure}

In~\cref{fig:image_prompts2}, we show image prompts corresponding to the image-conditioned multi-view generations in~\cref{fig:mv_supp}.

\end{document}